%% file: jin_477.tex
\documentclass[accepted]{uai2022} % after acceptance, for a revised
\usepackage[american]{babel}
\usepackage{natbib} % has a nice set of citation styles and commands
\usepackage{mathtools} % amsmath with fixes and additions
\usepackage{booktabs} % commands to create good-looking tables
\usepackage{tikz} % nice language for creating drawings and diagrams

\usepackage{tikz-cd}
\usepackage{multirow}
\usepackage{subfig}
\usepackage[colorlinks=true,linkcolor=red,citecolor=blue,urlcolor=black]{hyperref}
\input{Definitions}
\graphicspath{{./}}
\newcommand{\specialcell}[2][c]{%
  \begin{tabular}[#1]{@{}c@{}}#2\end{tabular}}
\title{Orthogonal Gromov-Wasserstein Discrepancy with Efficient Lower Bound}

\author[1]{\href{mailto:<hjin25@uic.edu>?Subject=Your UAI 2022 paper}{Hongwei Jin}{}}
\author[1]{Zishun Yu} 
\author[1]{Xinhua Zhang}
\affil[1]{%
    Computer Science Department\\
    University of Illinois at Chicago\\
    Chicago, Illinois, USA
}
  
\begin{document}
\maketitle

\begin{abstract}
  Comparing structured data from possibly different metric-measure spaces is a fundamental task in machine learning, with applications in,
  e.g., graph classification.
  The Gromov-Wasserstein (GW) discrepancy formulates a coupling between the structured data based on optimal transportation,
  tackling the incomparability between different structures by aligning the intra-relational geometries.
  Although efficient \emph{local} solvers such as conditional gradient and Sinkhorn are available,
  the inherent non-convexity still prevents a tractable evaluation,
  and the existing lower bounds are not tight enough for practical use.
  To address this issue, we take inspirations from the connection with the quadratic assignment problem,
  and propose the orthogonal Gromov-Wasserstein (OGW) discrepancy as a surrogate of GW.
  It admits an efficient and \emph{closed-form} lower bound with $\mathcal{O}(n^3)$ complexity,
  and directly extends to the fused Gromov-Wasserstein (FGW) distance, incorporating node features into the coupling.
  Extensive experiments on both the synthetic and real-world datasets show the tightness of our lower bounds, and both OGW and its lower bounds efficiently deliver accurate predictions and satisfactory barycenters for graph sets.
\end{abstract}

\section{Introduction}
 \label{sec:intro}
 Similarity based learning has been a popular approach in many machine learning applications.
 Instead of directly modeling each individual object which may pose challenge in some areas,
 it resorts to models based on pairwise similarity,
 possibly across different domains.
 The most common example is the kernels used in support vector machines and Gaussian process,
 including RBF kernel and covariance matrices that measure similarity in the Euclidean space \citep{chang2010training},
 and string or tree kernels that compare discrete objects \citep{LodSauShaCrietal02}.
 More generally, featured graphs have been a useful tool for capturing similarities and relations in the structured data that are commonly not Euclidean.
 Examples include social network \citep{fan2019graph},
 recommendation systems \citep{wu2020graph},
 fraud detection \citep{li2020flowscope},
 quantum chemistry \citep{coley2019graph}
 and topology-aware IoT applications \citep{abusnaina2019adversarial}.

 Despite the inherent challenge in comparing graphs from possibly different metric-measure spaces,
 % with the important notion of isometry .
 there has been a wealth of refined discrepancy measures between graphs,
 including kernels \citep{VisSchKonBor10,SheSchvanMehetal11} and GCNs based approaches \citep{BroBruLeCSzletal17,DefBreVan16}.
 Recently, the Gromov-Wasserstein discrepancy \citep[$\gw$,][]{PeyCutSol16},
 which extends the Gromov-Wasserstein distance \citep{Memoli11},
 has emerged as an effective transportation distance between structured data,
 alleviating the incomparability issue between different structures by aligning the \textit{intra}-relational geometries.
 Thanks to its favorable properties such as efficiency and isometry-awareness,
 $\gw$ has been applied to
 domain adaptation \citep{YanLiWuMinetal18},
 word embedding \citep{AlvJaa18},
 graph classification \citep{VayChaFlaTavetal19},
 metric alignment \citep{EzuSolKimBen17},
 generative modeling \citep{CohSej19},
 and graph matching and node embedding \citep{XuLuoZhaCar19,XuLuoCar19}.

 However, different from the standard Wasserstein distance which is a linear program,
 $\gw$ is unfortunately intractable to evaluate.
 Despite the practical success of non-convex optimization techniques such as conditional gradient method and entropic regularization \citep{PeyCutSol16, gold1996graduated},
 it remains NP-hard to find the global optimum.
 Hence, existing practice settles with local solutions, lacking an analyzable guarantee.
 This significantly challenges the trustworthiness of the $\gw$ discrepancy.

 Towards a tractable approximation, \citet{Memoli11} proposed three lower bounds of $\gw$, which cost $O(n^2)$, $O(n^4)$ and $O(n^5)$.
 They can be useful for branch-and-bound based global optimization,
 as well as recent algorithms for certifying the robustness of nonconvex models.
 However, these lower bounds are rarely used in practice,
 and as we will show later in Figure~\ref{fig:tightness},
 even the most expensive lower bound can be quite loose,
 raising the concerns on their effectiveness.

 Instead of developing yet another lower bound or tight approximation of $\gw$, our goal in this paper is to design a \textit{surrogate} of it (namely orthogonal $\gw$, or $\ogw$) such that:
 \begin{enumerate}[label=(\alph*)]
   \item
         It retains $\gw$'s desirable properties such as permutation invariance, non-negativity, triangle inequality, and good performance in machine learning tasks such as classification and barycenter.  We stress that $\ogw$ does not need to be either an upper bound or a lower bound of $\gw$.  We are also \emph{not} concerned about the gap between $\ogw$ and $\gw$ because what matters is the desirable mathematical properties and the performance in learning, instead of how close $\ogw$ is to $\gw$.
   \item
         It does \emph{not} have to be tractable, but it must admit a \emph{tight and efficient} lower bound  -- upper bound is easier from local optimization as in $\gw$.  The tightness will potentially contribute to global optimization such as certification, a topic that is beyond this paper.  Ideally, such a lower bound should also possess the aforementioned good properties of the surrogate itself.
 \end{enumerate}

 Our inspiration, as unrolled in Section \ref{sec:GW_div},
 stems from the connection of $\gw$ to quadratic assignment problems (QAPs),
 which was tapped into by sliced $\gw$ \citep{VayFlaTavChaetal19} and
 Gromov-Monge problems \citep{MemNee21}.
 It paved the way for approximating the set of doubly stochastic matrices (used in $\gw$) by orthonormal matrices under marginal constraints \citep{RenWol92,HadRenWol92,AnsBri01}.
 The resulting problem is well known to admit tight approximate solutions,
 and accommodates fused $\gw$ to account for node features \citep[$\fgw$,][]{VayChaFlaTavetal19}.
 Experiments on classification and barycenter demonstrate the effectiveness of our proposed $\ogw$ and its lower bounds.

\section{$\ogw$-Discrepancy with Tractable Lower Bounds}
 \label{sec:GW_div}

 We represent an undirected graph $G$ with $n$ nodes by an adjacency matrix
 $\Avec \in \{0,1\}^{n \times n}$,
 where $\Avec_{ii} = 0$,
 and $\Avec_{ij} = 1$ if there is an edge between nodes $i$ and $j$ ($i \neq j$),
 and $0$ otherwise.
 % REVIEW: remove the constraints of same size
 To start with, we consider the discrepancy between two graphs with the same order (\ie, number of nodes).
 A detailed generalization to any graph orders will be addressed in Section~\ref{subsec:graphs_diff_size}.
 Associated with the nodes is a distribution,
 encoding some prior information about their importance,
 e.g., the normalized degree of each node \citep{XuLuoZhaCar19}.
 However, many applications lack such natural normalization \citep{VayChaFlaTavetal19, VayFlaTavChaetal19, PeyCutSol16}.
 \iffalse
 % NOTE: Remove the non-uniform appendix
 \todo{keep it or not?}
 However, our experiments in Appendix~\ref{sec:exp_nonuniform} show that it makes no statistically significant improvement on prediction accuracy compared with a simple uniform distribution,
 which is also a popular choice such as in \citep{VayChaFlaTavetal19, VayFlaTavChaetal19, PeyCutSol16}.
 \todo{rephrase the non-uniform distribution is an extension from our formulation. REMOVED}
 We include our extension to any arbitrary node distribution in Appendix~\ref{sec:app_tight_QU}
 Indeed, many applications lack such natural normalizations.
 \fi
 Therefore, we will stick with a uniform distribution over all nodes and
 include our extension of non-uniform distribution in Appendix~\ref{subsec:app_node_dist}.
 Letting $\one = (1,\ldots,1)^\top$ whose dimensionality can be implicitly induced from the context,
 the standard $\gw$ distance between graph $G$ and $H$ based on $\ell_2$ distance can be formulated as (the square root of)
 \begin{align}
   \label{eq:def_gw}
    & \nonumber \gw(G, H) :=            \\
    & \min_{\Pvec \in \Ecal \cap \Ncal}
   \sum_{i,j,k,l} \sbr{c_G(i,j) - c_H(k,l)}^2 \Pvec_{ik} \Pvec_{jl},
 \end{align}
 where $\Ecal := \{ \Pvec \in \RR^{n \times n} : \Pvec \one = \Pvec^\top \one = \one \}$
 and $\Ncal := \RR_+^{n \times n}$ \citep{Memoli11}.
 Here $c_G(i,j)$ represents a distance measure between node $i$ and $j$ on $G$,
 and common choices include their shortest-path distance,
 or simply $1 - \Avec_{ij}$ (the complement of adjacency).
 When both $c_G$ and $c_H$ are a metric,
 \eqref{eq:def_gw} is a squared metric on isomorphism classes of measurable metric spaces.
 However, as pointed out by \citet{PeyCutSol16},
 $c_G$ and $c_H$ do not have to be restricted to metrics and $\ell_2$ can be extended to other asymmetric or non-subadditive losses such as $f$-divergence.
 They call it $\gw$ discrepancy,
 which broadens its applicability in machine learning.
 We will refer to it as $\gw$,
 without even taking the square root of \eqref{eq:def_gw} just like in \citet{PeyCutSol16}.

 \begin{remark}
   \label{rmk:similarity}
   It is noteworthy that although the original $\gw$ requires $c_G(i,j)$ to be a distance metric \citep{Memoli11},
   it can be relaxed in \eqref{eq:def_gw} where $\ell_2$ loss is used.
   Indeed, $c_G(i,j)$ can also be served by \emph{similarities} between nodes instead of distance, \eg, by simply flipping the sign of the distance measure. This also opens up the use of non-metric dissimilarity measures such as constrained shortest path \citep{LozMed13}.
 \end{remark}

 Define two $n$-by-$n$ symmetric matrices $\Cvec$ and $\Dvec$ whose $(i,j)$-th elements are $\Cvec_{ij} = c_G(i,j)$ and $\Dvec_{ij} = c_H(i,j)$, respectively.
 For example,
 the complement of adjacency can be written as $\Cvec = \one \one^\top - \Avec$.
 The above $\gw$ can be compactly rewritten in the Koopmans-Beckmann form \citep{KooBec57}:
 \begin{align}
   \label{eq:obj_qap}
    & \nonumber \gw(G,H) = \\
    & \frac{1}{n^2}\rbr{
     \nbr{\Cvec}_F^2
     + \nbr{\Dvec}_F^2
     - 2 \max_{\Pvec \in \Ecal \cap \Ncal}
     \tr(\Cvec \Pvec \Dvec \Pvec^\top)}.
 \end{align}
 Here $\nbr{\cdot}_F$ is the Frobenius norm.
 Obviously,
 $\gw$ is permutation invariant,
 nonnegative,
 and equals 0 when $G$ and $H$ are isomorphic.
 The major drawback is that the maximization over $\Pvec$ is intractable,
 although efficient \textit{local} algorithms are available such as conditional gradient \citep{VayChaFlaTavetal19} and Sinkhorn \citep{PeyCutSol16}.

 \subsection{Connecting $\ogw$ with the quadratic assignment problem}
  \label{subsec:conn_qap}

  Rewriting $\gw$ with quadratic optimization over $\Ecal \cap \Ncal$ as in \eqref{eq:obj_qap} reveals an innate connection to the quadratic assignment problem (QAP).
  Noting that by the Birkhoff–von Neumann theorem \citep{Birkhoff46},
  $\Ecal \cap \Ncal$ is the convex hull of the set of $n \times n$ permutation matrices (denoted as $\Pi$).
  % That is $\Pcal = \text{ext}(\Ecal \cap \Ncal)$.
  Indeed, the connection with QAP has been used to formulate Gromov-Monge distances \citep{MemNee21},
  and to accelerate the evaluation of $\gw$ via projection (slicing) to 1-D \citep{VayFlaTavChaetal19}.
  Fortunately, a number of tractable relaxations of QAP are available,
  many of which are based on the following characterization of $\Pi$:
  \begin{align}
    \Pi = \Ecal \cap \Ncal \cap \Ocal,
  \end{align}
  where $\Ocal := \{ \Pvec \in \RR^{n \times n} : \Pvec^\top \Pvec = \Pvec \Pvec^\top = \Ivec\}$.
  Here $\Ivec$ is the identity matrix.
  Whenever necessary,
  we will explicitize the dimensionality of $\Ocal$ by writing $\Ocal_n$.
  Interestingly, $\tr(\Cvec\Pvec\Dvec\Pvec^\top)$ can be maximized exactly by a simple eigen-decomposition if $\Pvec$ is restricted to $\Ocal$ \citep{Umeyama88}.
  Specifically,
  assume the eigen-decomposition of $\Cvec$ and $\Dvec$ are $\Cvec = \Pvec_\Cvec \diag(\lambdavec_\Cvec) \Pvec_\Cvec^\top$ and $\Dvec = \Pvec_\Dvec \diag(\lambdavec_\Dvec) \Pvec_\Dvec^\top$, respectively,
  and suppose the eigenvalues in $\lambdavec_\Cvec$ and $\lambdavec_\Dvec$ are both arranged in a descending order.
  Then
  \begin{align}
    \label{eq:opt_Q_quad}
     & \Pvec_1 \Pvec_2^\top
    \in
    \argmax_{\Pvec \in \Ocal}\ \tr(\Cvec\Pvec\Dvec\Pvec^\top),
    \quad                                                                   \\
    \text{and} \quad
     & \max_{\Pvec \in {\color{red} \Ocal}}\ \tr(\Cvec\Pvec\Dvec\Pvec^\top)
    = \lambdavec_\Cvec^\top \lambdavec_\Dvec.
  \end{align}
  Based on this result, \citet{HadRenWol92} proposed tightening the domain approximation from $\Ocal$ to $\Ocal \cap \Ecal$,
  which, despite the original intention of approximating inhomogeneous QAPs,
  happens to be useful in our context too.
  Compared with $\Ncal \cap \Ecal$, $\Ocal \cap \Ecal$ offers more convenience in constructing upper and lower bounds that are tight and efficient.
  This substitution leads to our proposed new metric,
  named as \emph{orthogonal Gromov-Wasserstein ($\ogw$)} discrepancy:
  \begin{align}
    \label{eq:obj_gwtil}
     & \nonumber \ogw(G,H) :=                               \\
     & \frac{1}{n^2} \rbr{\nbr{\Cvec}_F^2 + \nbr{\Dvec}_F^2
      - 2 \max_{\Pvec \in {\color{red}\Ocal \cap \Ecal}} \tr(\Cvec \Pvec \Dvec \Pvec^\top)}.
  \end{align}
  Figure~\ref{fig:conn_qap_gw_ogw} illustrates how QAP is connected with $\gw$ and $\ogw$ through the different convex outer approximations of the domain of the permutation matrices.
  \input{fig_qap_connection.tex}

 \subsection{Upper and lower bounds of $\ogw$}
  \label{subsec:ub_lb_ogw}

  The evaluation of $\ogw$ is hindered by the nonconvex objective \emph{and} the nonconvex domain in the optimization of $\Pvec$ in \eqref{eq:obj_gwtil}.
  So it is natural to resort to its lower and upper bounds.

  \paragraph{Upper bound of $\ogw$.}

  Obviously, any locally optimal $\Pvec$ in \eqref{eq:obj_gwtil} yields an upper bound of $\ogw$.
  To ease the local optimization, we first leverage the characterization of $\Ocal \cap \Ecal$  \citep{HadRenWol92}:
  \begin{align}
    \label{eq:OE_char}
    \Ocal \cap \Ecal
    =
    \cbr{\smallfrac{1}{n} \one \one^\top + \Vvec \Qvec \Vvec^\top
      :  \Qvec \in \Ocal_{n-1}},
  \end{align}
  where $\Vvec$ is any $n \times (n-1)$ matrix satisfying $\Vvec^\top \one = \zero$ and $\Vvec^\top \Vvec = \Ivec_{n-1}$.
  An example is given in Appendix~\ref{sec:Vmatrix}.
  Plugging $\Pvec = \smallfrac{1}{n} \one \one^\top + \Vvec \Qvec \Vvec^\top$ into the optimization objective in \eqref{eq:obj_gwtil} yields
  \begin{align}
    \label{eq:def_gw_Q}
     & \nonumber \max_{\Pvec \in \Ocal \cap \Ecal} \tr(\Cvec \Pvec \Dvec \Pvec^\top)
    =                                                                                \\
     & \frac{1}{n^2} s_{\Cvec} s_{\Dvec}
    + \underbrace{
      \max_{\Qvec \in \Ocal} \ \{\tr (\vec\Chat \Qvec \vec\Dhat \Qvec^\top)
      + \tr (\vec\Ehat^\top \Qvec)\}}_{=: \ {\color{red} \Qcal(\vec\Chat, \vec\Dhat, \vec\Ehat)}},
  \end{align}
  % 
  % \todo{replace $\Qcal$ along with the const term}
  % \begin{align}
  %    & \Qcal(\Cvec, \Dvec)
  %   = \max_{\Qvec \in \Ocal} \tr (\vec\Chat \Qvec \vec\Dhat \Qvec^\top)
  %   +\tr (\vec\Ehat^\top \Qvec)
  %   +\smallfrac{1}{n^2} s_{\Cvec} s_{\Dvec}
  % \end{align}
  % 
  where $\vec\Xhat := \Vvec^\top \Xvec \Vvec$ and $s_{\Xvec} := \one^\top \Xvec \one$
  for any matrix $\Xvec$, and $\Evec := \smallfrac{2}{n} \Cvec \one \one^\top \Dvec$.
  Since $\Qcal(\vec\Chat, \vec\Dhat, \vec\Ehat)$ involves both linear and quadratic terms in $\Qvec$,
  no closed-form solution remains available.
  %   This motivates us to develop lower and upper bounds for $\Qcal$,
  %   which will serve as a tractable solution to the $\ogw$ discrepancy under isometry.
  % which will serve as pillars of attacking and certification algorithms, respectively.

  Clearly, any locally optimal $\Qvec$ yields a \emph{lower bound} for $\Qcal$ (denoted as $\Qcal_{lb}$),
  \ie, an upper bound for $\ogw$ (denoted as $\ogw_{ub}$).
  \textit{Locally} optimizing $\Qvec$ over $\Ocal$ (\emph{a.k.a.} Stiefel manifold) has been very well studied \citep{AbsMahSep09,WenYin13,AraMoh18},
  and we adopt a straightforward approach of projected quasi-Newton,
  noting that the projection of any matrix $\Qvec$ on $\Ocal$ is simply $\Uvec_{\Qvec} \Vvec_{\Qvec}^\top$, where the singular value decomposition (SVD) of $\Qvec$ is $\Uvec_{\Qvec} \Lambda_{\Qvec} \Vvec_{\Qvec}^\top$.
  With the locally optimal $\Qvec$ in hand,
  the locally optimal $\Pvec$ for $\ogw$ can be recovered by plugging $\Qvec$ into the formula in \eqref{eq:OE_char}.

  Similarly to the practice of $\gw$ which resorts to locally optimal solutions,
  we will use $\ogw_{ub}$ as a practical ``evaluation'' of $\ogw$.
  Whenever there is no confusion (especially in empirical investigation),
  we will simply refer to the performance of $\ogw_{ub}$ as the performance of $\ogw$.

  \paragraph{Lower bounds of $\ogw$.}
  The simplest way to lower bound $\ogw$ is by relaxing the domain of $\Pvec$ into $\Ocal$ in \eqref{eq:obj_gwtil}:
  \begin{align}
    \label{eq:obj_ogw_o}
          & \nonumber \ogw_o(G,H)                                           \\
    := \  & \frac{1}{n^2}\rbr{
      \nbr{\Cvec}_F^2
      + \nbr{\Dvec}_F^2
      - 2 \max_{\Pvec \in \Ocal}
    \tr(\Cvec \Pvec \Dvec \Pvec^\top)}                                      \\
    = \   & \smallfrac{1}{n^2} \nbr{\lambdavec_\Cvec - \lambdavec_\Dvec}^2,
  \end{align}
  where the last step is by \eqref{eq:opt_Q_quad}.
  We note in passing that $\ogw_o$ embodies a different design principle from heat kernel signature \citep{SunOvsGui09} and wave kernel signature \citep{AubSchCre11},
  in that neither of the kernel signatures sort the kernel spectrum.

  In practice, we found that completely dropping the constraint $\Ecal$ may lead to over relaxation.
  To bring back $\Ecal$, we follow \citet{HadRenWol92} and decompose $\Qcal$ in \eqref{eq:def_gw_Q} into quadratic and linear terms by decoupling their $\Qvec$:
  \begin{align}
    \label{eq:def_qcal_ub}
     & \nonumber \Qcal_{ub}(\vec\Chat, \vec\Dhat, \vec\Ehat)                      \\
     & := \max_{\Qvec_1 \in \Ocal} \tr (\vec\Chat \Qvec_1 \vec\Dhat \Qvec_1^\top)
    + \max_{\Qvec_2 \in \Ocal} \tr (\vec\Ehat^\top \Qvec_2).
  \end{align}
  As a result, we obtain an \emph{upper bound} of $\Qcal$ (denoted as $\Qcal_{ub}$),
  which produces a \emph{lower bound} of $\ogw$:
  \begin{align}
    \label{eq:ogw_lb}
     & \ogw_{lb}(G, H) :=                                   \\
    \nonumber
     & \frac{1}{n^2} \rbr{\nbr{\Cvec}_F^2 + \nbr{\Dvec}_F^2
    - 2 \, \Qcal_{ub}(\vec\Chat, \vec\Dhat, \vec\Ehat)
    -  \frac{1}{n^2} s_{\Cvec}s_{\Dvec}}.
  \end{align}
  $\Qcal_{ub}$ can be evaluated analytically.
  First, $\Qvec_1$ can be solved by \eqref{eq:opt_Q_quad}.
  As for $\Qvec_2$, the von Neumann's trace inequality implies that its optimal value is
  $\Uvec_{\Evec} \Vvec_{\Evec}^\top$,
  where $\Uvec_{\Evec} \vec\Lambda_{\Evec} \Vvec_\Evec^\top$ is the SVD of $\vec\Ehat$,
  and the maximum value of $\tr(\vec\Ehat^\top \Qvec_2)$ is $||\vec\Ehat||_*$,
  the trace norm of $\vec\Ehat$,
  which is the sum of the singular values of $\vec\Ehat$.
  % except for special cases such as $\one$ being an eigenvector of $C$ or $D$,
  \citet{HadRenWol92} showed that such an upper bound in \eqref{eq:def_qcal_ub} is often quite tight,
  which is also observed in our experiments.
  Indeed, we noticed that the magnitude of $\vec\Chat$ and $\vec\Dhat$
  in \eqref{eq:def_qcal_ub} is significantly larger than that of $\vec\Ehat$.
  Therefore, although the optimal $\Qvec_1$
  and $\Qvec_2$ are different, the resulting gap is small.

  We next summarize the mathematical properties of $\ogw$ and its lower bounds as follows:
  \begin{theorem}
    \label{thm:ogw}
    $\ogw$, $\ogw_o$, and $\ogw_{lb}$ are all nonnegative and symmetric.
    Their square root satisfies the triangle inequality.
    Their values are 0 if (but not only if) the two graphs are isomorphic.
  \end{theorem}
  The proof is in Appendix~\ref{sec:proof_nonneg_ogw}.
  Compared with the requirement of distance metric,
  $\ogw$ and its lower bounds only fall short of the ``only if'' part of the identity of indiscernibles.
  To see why ``only if''cannot hold,
  consider $\ogw_o$ whose closed form in \eqref{eq:obj_ogw_o} shows that its value can be $0$ as long as $\Cvec$ and $\Dvec$ are similar,
  \ie, share the same eigenvalues.
  In general, however, $\Cvec$ and $\Dvec$ are derived from graphs with certain \textit{discrete} properties,
  leaving permutation the most likely path to similarity.
  %   The condition for $\ogw$ and $\ogw_{lb}$ to be zero is even more stringent.
  \begin{remark}
    The coupling matrix $\Pvec$ in $\gw$ provides a useful matching between two sets of nodes. Although the $\Pvec$ in $\ogw$ and its lower bounds may contain negative entries, it optimizes over the orthonormal domain,
    which may still provide useful insights between the two groups of node.
    For example, invariance to orthogonal transformation is a longstanding pursuit in learning \citep{KorNorLeeHin19}.
    Despite the hardness of exactly optimizing $\Pvec$ for $\ogw$, we can use $\Qvec_1$ from \eqref{eq:def_qcal_ub} to recover $\Pvec$ via the transformation in \eqref{eq:OE_char}.
    This is reasonable because $\hat\Evec$
    is generally much smaller in magnitude than $\hat\Cvec$ and $\hat\Dvec$.
  \end{remark}

  %  \subsection{Computation complexity}
  % \label{subsec:complexity}
  % The convergence rate of projected quasi-Newton method determines the convergence rate of $\ogw_{ub}$ \todo{state the rate}.
  % https://www.cs.utexas.edu/users/inderjit/public_papers/pqnj_sisc10.pdf
  \paragraph{Computational complexity.}
  The analytic solution for $\ogw_{lb}$ and $\ogw_o$ is achieved by singular value decomposition and eigen decomposition, whose computational complexity is $\Ocal(n^3)$.
  $\vec\Chat$, $\vec\Dhat$, and $\vec\Ehat$ can be computed in $\Ocal(n^3)$ thanks to the structure of $\Vvec$ (see Appendix \ref{sec:Vmatrix}).
  % , while the best complexity recently is $\Ocal(n^{2.3728596})$ \citep{alman2021refined}.
  % https://en.wikipedia.org/wiki/Computational_complexity_of_matrix_multiplication
  % To simplify, we implement matrix operations in a conventional way.
  It is worth mentioning that \citet{Memoli11} also derived the lower bounds of $\gw$ by solving a set of linear assignment problems,
  named First Lower Bound (FLB), Second Lower Bound (SLB), and Third Lower Bound (TLB).
  The following inequalities provide the connections between different lower bounds of $\gw$:
  \begin{align}
    \gw \ge \begin{cases}
              \gw_{tlb} \ge \gw_{flb} \\
              \gw_{slb}.
            \end{cases}
  \end{align}
  And the complexities of FLB, SLB and TLB are $\Ocal(n^2), \Ocal(n^4), \Ocal(n^5)$, respectively.
  In general, TLB provides the tightest lower bound for $\gw$, and SLB has resemblant performance compared with TLB.
  This is also observed in our experiments.

 \subsection{Graphs with Different Sizes}
  \label{subsec:graphs_diff_size}

  So far, we have been restricting the two graphs to have the same number of nodes,
  primarily because the orthonormal domain $\Ocal$ only contains square matrices.
  In order to deal with the non-square matrix, \ie, graphs of different order,
  we introduce the semi-orthogonal domain
  \begin{align}
    \tilde{\Ocal}_{m, n} := \{ \Tvec \in \RR^{m \times n}: \Tvec^\top \Tvec= \Ivec_n \},
  \end{align}
  where $m > n$ without loss of generality.
  That is a domain of ``tall'' matrices, whose columns are orthonormal.
  When $m=n$, $\tilde{\Ocal}$ recovers $\Ocal_n$ because $\Tvec^\top \Tvec= \Ivec_n$ is equivalent to $\Tvec \Tvec^\top = \Ivec_n$ for a square matrix $\Tvec$.

  For a non-square matrix $\Pvec \in \RR^{m \times n}$ and $m > n$,
  it is no longer feasible to impose the constraint of
  $\Ecal := \{ \Pvec \in \RR^{n \times n} : \Pvec \one = \Pvec^\top \one = \one \}$ because the dimensionality does not match.
  Instead, it can be generalized into
  \begin{align}
    \tilde{\Ecal} := \{
    \Pvec \one_n = \sqrt{\smallfrac{n}{m}}\one_m,
    \Pvec^\top \one_m = \sqrt{\smallfrac{m}{n}}\one_n
    \}.
  \end{align}
  To summarize,
  we can extend the $\ogw$ in \eqref{eq:obj_gwtil} to graphs with different orders by replacing the domain of $\Pvec$, amounting to
  \begin{align}
    \label{eq:obj_gwtil_nonsquare}
              & \ogw(G,H) :=                                                  \\
    \nonumber & \frac{1}{m^2} \nbr{\Cvec}_F^2 + \frac{1}{n^2} \nbr{\Dvec}_F^2
    - \frac{2}{mn} \max_{\Pvec \in {\color{red}\tilde{\Ocal} \cap \tilde{\Ecal}}}
    \tr(\Cvec \Pvec \Dvec \Pvec^\top),
  \end{align}
  where $G$ and $H$ have the graph order of $m$ and $n$, respectively.
  We reuse the symbol $\ogw$ because the constraints $\tilde{\Ocal}$ and $\tilde{\Ecal}$ recover $\Ocal$ and $\Ecal$ respectively when $m = n$.
  It is also easy to see that $\ogw$ is nonnegative because
  \begin{align*}
        & \ogw(G,H)                                                                            \\
    \ge & \frac{1}{m^2} \nbr{\Cvec}_F^2 + \frac{1}{n^2} \nbr{\Dvec}_F^2
    - \frac{2}{mn} \max_{\Pvec \in {\color{red}\tilde{\Ocal} }}
    \tr(\Cvec \Pvec \Dvec \Pvec^\top)                                                          \\
    =   & \nbr{\smallfrac{1}{m} \lambdavec_\Cvec - \smallfrac{1}{n} \lambdavec_\Dvec}^2 \ge 0.
  \end{align*}
  In the similar spirit to \eqref{eq:OE_char},
  any $\Pvec \in \tilde{\Ocal} \cap \tilde{\Ecal}$ can be re-parameterized as follows
  \begin{theorem}
    \label{thm:nonsquare_reparam}
    Let $\Uvec \in \RR^{m \times (m-1)}$ and
    $\Vvec \in \RR^{n \times (n-1)}$ be arbitrary projection matrices satisfying
    \begin{align}
      \Uvec^\top \one_m & = \zero_{m-1},\  \Vvec^\top \one_n                = \zero_{n-1},\ \\
      \Uvec^\top \Uvec  & = \Ivec_{m-1},\  \ \Vvec^\top \Vvec = \Ivec_{n-1}.
    \end{align}
    Then
    \begin{align}
      \label{eq:P_param_nonsquare}
      \Pvec \in \tilde{\Ocal} \cap \tilde{\Ecal}
      \ \ \iff \ \
      \Pvec  = \smallfrac{1}{\sqrt{mn}} \one_m \one_n^\top + \Uvec \Qvec \Vvec^\top,
    \end{align}
    where $\Qvec \in \tilde{\Ocal}_{m-1,n-1}$.
  \end{theorem}

  The proof is relegated to Appendix~\ref{subsec:app_p_param_nonsquare}.
  Letting $\vec\Chat := \Uvec^\top \Cvec \Uvec$ and $\vec\Dhat := \Vvec^\top \Dvec \Vvec$,
  the optimization over $\Pvec$ in \eqref{eq:obj_gwtil_nonsquare} turns into a projected QAP (PQAP) with an additional linear term and some constant terms
  \begin{align}
    \max_{\Qvec \in \tilde{\Ocal}_{m-1,n-1}}\ \tr(\vec\Chat \Qvec \vec\Dhat \Qvec^\top) + \tr(\vec\Ehat \Qvec^\top) + \text{const},
  \end{align}
  where $\vec\Ehat = \smallfrac{2}{\sqrt{mn}} \Uvec^\top \Cvec \one_m \one_n^\top \Dvec \Vvec$.
  Finally, in order to leverage the favorable properties of orthonormal matrices,
  we right-pad the matrix $\Qvec$ by an $(m-1)\times (n-m)$ matrix $\Jvec$,
  such that $\sbr{\Qvec, \ \Jvec} \in \Ocal_{m-1}$.
  Indeed
  \begin{align}
    \label{eq:padding-projection}
    \Qvec \in \tilde{\Ocal}_{(m-1), (n-1)} \ \iff \
    \exists \Jvec:
    \begin{bmatrix} \Qvec & \Jvec \end{bmatrix} \in \Ocal_{m-1}.
  \end{align}
  Such a $\Jvec$ matrix only needs to be any basis of the kernel space of $\Qvec$.
  As a result, the quadratic term
  $\max_{\Qvec \in \tilde{\Ocal}}\ \tr(\vec\Chat \Qvec \vec\Dhat \Qvec^\top)$
  is equivalent to
  \begin{align}
    \label{eq:padding_quad}
    % \max_{\begin{bmatrix} \Qvec & \Xvec \end{bmatrix} \in \Ocal}
    \max_{\sbr{\Qvec\ \Jvec} \in \Ocal}
    \tr\rbr{ \vec\Chat \begin{bmatrix} \Qvec & \Jvec \end{bmatrix}
      \begin{bmatrix}
        \vec\Dhat & \zero \\
        \zero     & \zero
      \end{bmatrix}
      \begin{bmatrix}
        \Qvec & \Jvec
      \end{bmatrix}^\top},
  \end{align}
  and we can optimize $\sbr{\Qvec\ \Jvec}$ as a whole,
  utilizing the closed-form solution to the squared matrix case.
  \iffalse
  where the columns in $\Jvec$ are orthogonal to the columns in $\Qvec$,
  \ie, $\Qvec^\top \Jvec = \zero$.
  \begin{align}
    \begin{bmatrix}\Qvec^\top \\ \T{\Jvec}\end{bmatrix}
    \begin{bmatrix}\Qvec & \Jvec\end{bmatrix}
     & = \begin{bmatrix}
           \Qvec^\top \Qvec & \zero           \\
           \zero            & \T{\Jvec} \Jvec
         \end{bmatrix}
    = \Ivec_{m-1}.
  \end{align}
  \fi

  Similarly, for the linear term, we have
  \begin{align}
    \label{eq:padding_linear}
    \max_{\Qvec \in \tilde{\Ocal}} \ \tr(\vec\Ehat \Qvec^\top) & =
    % \max_{\begin{bmatrix}\Qvec & \Kvec\end{bmatrix} \in \Ocal}
    \max_{\sbr{\Qvec\ \Kvec} \in \Ocal}
    \tr \rbr{
      \begin{bmatrix}
        \vec\Ehat & \mathbf{0}
      \end{bmatrix}
      \begin{bmatrix}
        \Qvec & \Kvec
      \end{bmatrix}^\top
    },\!\!
  \end{align}
  where $\Kvec$ serves the same role as $\Jvec$.
  % REVIEW: 
  % for the linear term, the original problem can be identified as 
  % ``orthogonal Procrusters problem" (https://en.wikipedia.org/wiki/Orthogonal_Procrustes_problem)
  % It is not necessary to be orthonormal, but orthogonal of columns
  % Once again, we have
  % \begin{equation}
  %   \Qvec \in \tilde{\Ocal}_{(m-1) \times (n-1)} 
  %   \iff 
  %   \exists \Kvec: 
  %   \begin{bmatrix}\Qvec & \Kvec\end{bmatrix} \in \Ocal_{m-1},
  % \end{equation}
  % where the columns in $\Kvec$ are orthogonal to the columns in $\Qvec$, 
  % \ie, $\Qvec^\top \Kvec = \zero$.
  % To see that, recall that by taking the SVD of $\sbr{\hat{\Evec}\ \zero}$, we will get the solution of $\sbr{\Qvec\ \Kvec}$ in the orthogonal domain.
  %
  % \begin{align}
  %   \begin{bmatrix}\vec\Ehat & \mathbf{0}\end{bmatrix}_{m \times m}
  %    & = \uvec \Sigma \T{\vvec}                          \\
  %   \begin{bmatrix}\Qvec & \Yvec\end{bmatrix}_{m \times m}
  %    & = \uvec \T{\vvec}                                 \\
  %   \begin{bmatrix}\T{\Qvec} \\ \T{\Yvec}\end{bmatrix}
  %   \begin{bmatrix}\Qvec & \Yvec \end{bmatrix}
  %    & = \begin{bmatrix}
  %          \T{\Qvec} \Qvec & \T{\Qvec} \Yvec \\
  %          \T{\Yvec} \Qvec & \T{\Yvec} \Yvec
  %        \end{bmatrix}               \\
  %    & = \uvec \T{\vvec} \vvec \T{\uvec}     = \Ivec_{m}
  % \end{align}

  To conclude,
  by padding zero on the non-square matrices $\vec\Dhat$ and $\vec{\Ehat}$ to square matrices,
  we can enjoy the analytic solutions to the problems in \eqref{eq:padding_quad} and \eqref{eq:padding_linear} in the same way as in the square case.

  \begin{remark}
    \label{rmk:padding}
    We refrain from the interpretation of adding dummy nodes with 0 distance because,
    as pointed out in Remark \ref{rmk:similarity},
    $\Cvec$ and $\Dvec$ can represent similarity measures.
    In such cases, padding with 0 is still justified with the above derivation,
    but not amenable to dummy node interpretations.
  \end{remark}

  \begin{remark}
    \label{rmk:disconnected}
    Disconnected graphs can be modeled by any existing heuristic that is also required by $\gw$. In \eqref{eq:def_gw},
    $c_G(i,j)$ cannot be $\infty$ because it would push $\gw$ to $\infty$ as long as all $c_H(k,l) < \infty$ and all nodes have nonzero marginals. A simple heuristic is to employ a large distance value between two nodes that belong to two separate/disconnected sub-graphs. Our experiments only involved connected graphs, because all the graphs from the real datasets are already connected -- none was discarded.
  \end{remark}

  We stress that the heuristic in Remark \ref{rmk:padding} is independent of padding 0 in \eqref{eq:padding_quad} that tackles different graph sizes. The latter is on how to characterize the alignment of two matrices, which is orthogonal to the design of base measure itself within each graph.

  %   \textit{Remark.}
  %   We humbly believe that it makes good sense to pad with 0.  Firstly, it is rigorously *derived* from the extension of the orthonormal matrix set $O$ to non-square matrices $O_{m,n}$ in Eq 12, where padding was not intended.
  %   Secondly, both $C$ and $D$ can represent similarity measures as in Peyre et al., (2016); see above point [c].  So we would like to develop a generic recipe.  Most importantly, $\ogw$ simply seeks the best alignment between $C$ and $D$ via the transformation matrix $P$.  To see the intuition of padding, let $C$ be a $2 \times 2$ matrix with eigenvalues $(\lambda_1, \lambda_2)$.  Then how to compare it with $D$, a $4 \times 4$ matrix whose eigenvalues form a vector $\mu \in \mathbb{R}^4$?  $\ogw_o$ simply pads $(\lambda_1, \lambda_2)$ with (0, 0) to make a 4-dimensional vector, which can be compared with $\mu$ via Euclidean distance.  We believe this makes good sense, and such a spectral padding is equivalent to padding zero matrix entries.  $\ogw$ and $\ogw_{lb}$ share the similar intuition (in a projected subspace), and their good performance has been corroborated by experimental results.

\section{Barycenter}
 \label{subsec:barycenter}

 We next study the application of $\ogw$ to the Barycenter problem,
 where given a set of sampled graphs and their associated weights,
 we aim to find their Fr\'echet mean by minimizing the weighted average discrepancy between the barycenter and the sampled graphs.
 Here the discrepancy is measured by our proposed $\ogw$,
 and the sampled graphs are represented by a set of cost matrices $\Dvec_i$,
 along with normalized weight $\lambda_i$ for $i \in \cbr{1,..., S}$.
 $\Dvec_i$ can also encode pairwise similarities instead of dissimilarities,
 and our method accommodates both cases naturally.
 The barycenter problem can be formalized as minimizing the weighted average $\ogw$ discrepancy:
 \begin{align}
   \label{eq:def_barycenter}
    & \min_{\Cvec} \Bcal(\Cvec)
   := \min \sum_{i=1}^S \lambda_i \cdot \ogw(\Cvec, \Dvec_i) \\
    & \nonumber
   = \sum_{i=1}^S \lambda_i
   \rbr{\frac{1}{m^2} \nbr{\Cvec}_F^2 - \frac{2}{m n_i} \max_{\Pvec_i \in \Ocal \cap \Ecal}\ \tr(\Cvec \Pvec_i \Dvec_i \Pvec_i^\top)}
   \\
    & \quad + \text{const}.
 \end{align}
 %
 %$\Pvec_i \in \tilde{\Ocal} \cap \tilde{\Ecal}$ is the transformation matrix between the barycenter and the $i$-th example.
 For simplicity, we specify the barycenter with a fixed order $m$,
 although the value of $m$ can also be optimized.

 We follow the block coordinate update proposed in \citet{PeyCutSol16},
 \ie, iteratively minimizing with respect to the couplings $\Pvec_i$ and updating the optimal cost matrix for the barycenter in a closed-form solution.
 Given a set of coupling matrices $\Pvec_i$,
 we can retrieve the optimal $\Cvec$ by taking the partial derivative of \eqref{eq:def_barycenter}
 \begin{align}
   \label{eq:bary_grad_C}
   \frac{\partial \Bcal(\Cvec)}{\partial \Cvec}
    & = \sum_i \lambda_i \rbr{\frac{2}{m^2} \Cvec
       - \frac{2}{mn_i} \Pvec_i \Dvec_i \Pvec_i^\top}
   \\
    & = \frac{2}{m^2} \Cvec
     - \sum_i \frac{2\lambda_i}{m n_i}\Pvec_i \Dvec_i \Pvec_i^\top
     = \zero.
 \end{align}
 So the optimal $\Cvec$ under the current $\{\Pvec_i\}$ is $\Cvec^*  \leftarrow  m  \sum_{i=1}^S \frac{\lambda_i}{n_i}\Pvec_i \Dvec_i \Pvec_i^\top$.
 Noting that $\ogw$ itself is still intractable,
 we resort to the tractable lower or upper bounds,
 replacing $\ogw$ in the definition of $\Bcal(\Cvec)$ with its tractable bounds.
 In the case of $\ogw_{ub}$ (resp. $\ogw_o$),
 we simply adopt the locally (resp. globally) optimal $\{\Pvec_i\}$.
 For $\ogw_{lb}$, we can rewrite $\Bcal(\Cvec)$ in terms of $\Qvec_1$ and $\Qvec_2$ from \eqref{eq:def_qcal_ub} and \eqref{eq:ogw_lb},
 and then $\Cvec$ can be updated using the optimal $\Qvec_1$ and $\Qvec_2$.
 More details are provided in Appendix~\ref{sec:grad_barycenter_ogw_lb}.

 It is noteworthy that any optimal solution $\Cvec$ for \eqref{eq:def_barycenter} leads to a set of optimal solutions $\{ \Pvec \Cvec \Pvec^\top : \Pvec \in \Ocal \cap \Ecal\}$.
 This also resonates with the intuition retrievable from $\ogw_o$ in \eqref{eq:obj_ogw_o}, where only the eigenvalues of $\Cvec$ matter.
 Therefore, additional ``post-processing'' is needed to pinpoint the optimal $\Cvec$ from the equivalent class.
 Furthermore, the optimal $\Cvec^*$ found over $\Ocal \cap \Ecal$ does not guarantee elementwise non-negativity.
 Next, we present our method, named as \emph{spectral reconstruction}, to find the appropriate $\Cvec$.

 \input{fig_transition.tex}
 \subsection{Spectral reconstruction}
  \label{subsec:eigen_proj}

  %   Since the $\ogw$ is defined on the domain of $\Ocal \cap \Ecal$ and optimized over $\Ocal$,
  %   we want to build a correspondence between $\Cvec^*$ with its domain properties.

  To begin with, suppose $\Cvec$ and all $\Dvec_i$ are graphs of order $m$,
  and we consider their projections to $\RR^{(m-1) \times (m-1)}$ via $\hat{\Cvec}^* = \Vvec^\top \Cvec^* \Vvec$ and $\hat{\Dvec}_i = \Vvec^\top \Dvec_i \Vvec$.
  Let their eigen-decomposition be
  $\hat{\Cvec}^* = R^\top \Sigma R$ and
  $\hat{\Dvec}_i = S_i^\top \Delta_i S_i$.
  Inspired by the above observation that $\ogw(\Cvec, \Dvec_i)$ depends primarily on the eigenvalues of $\Cvec$ and $\Dvec_i$,
  we rebuild $\hat{\Cvec}^*$ by $\hat{\Cvec}^*_{recon} = \sum_i \lambda_i S_i^\top \Sigma S_i$,
  \ie, trusting and retaining the eigenvalues of the optimal solution $\hat{\Cvec}^*$ while pairing them with the eigenvectors of the sampled graphs.

  For graphs with different sizes,
  the trick in Section~\ref{subsec:graphs_diff_size}, \ie, padding on smaller graph,
  helps us to assemble the $\vec\Chat^*$ from the top $m-1$ eigen system.
  Noting that $\hat{\Cvec}^*$ is still on the projected domain $\RR^{m-1 \times m-1}$, \ie,
  there exist $\Vvec$ such that
  $\hat{\Cvec}^* = \Vvec^\top \Cvec^* \Vvec$.

  % \todo{verify the claim}
  %   \begin{remark}
  %     \label{thm:same_eig}
  %     Given two graphs $G$ and $H$ with same order, and their corresponding cost matrices $\Cvec$ and $\Dvec$.
  %     The semi-orthogonal projection matrix $\hat{\Cvec}$ and $\hat{\Dvec}$, share the same set of eigenvalues.
  %   \end{remark}
  % To proof it, we have $\Cvec = \vec\Pi^\top \Dvec \vec\Pi$, then
  % $\vec\Chat = \Uvec^\top \vec\Pi^\top \Dvec \vec\Pi \Uvec = \vec\Pi^\top \Uvec^\top \Dvec \Uvec \vec\Pi = \vec\Pi^\top \vec\Dhat \vec\Pi$.

  Next, we bring $\hat{\Cvec}^*_{recon}$ back to its original space $\RR^{m \times m}$ via
  \begin{align}
    \Cvec^*_{recon} = \Vvec \hat{\Cvec}^*_{recon} \Vvec^\top + \Yvec,
  \end{align}
  % 
  % where $\Yvec \in \cbr{\Vvec^\top \Yvec \Vvec = \zero,  \diag(\Vvec \hat{\Cvec} \Vvec^\top + \Yvec) = \zero}$.
  where $\Yvec$ satisfies $\Vvec^\top \Yvec \Vvec = \zero$,
  ensuring that $\Vvec^\top \Cvec^*_{recon} \Vvec$ recovers $\hat{\Cvec}^*_{recon}$.
  In addition, when $\ogw$ operates on dissimilarity matrices,
  we require $\diag(\Cvec^*_{recon}) = \zero$, i.e.,
  the dissimilarity between a node and itself is 0.
  A straightforward choice of $\Yvec$ satisfying the two conditions is
  \begin{align}
    \label{eq:Y_mat}
     & \Yvec = - \frac{1}{2}(\dvec \one^\top + \one \dvec^\top),
    \\
    \where
     & \dvec := \diag(\Vvec \hat{\Cvec}^*_{recon} \Vvec^\top).
  \end{align}
  % 
  %   Trivially, it satisfies the $\diag(\Vvec \hat{\Cvec}^* \Vvec^\top + \Yvec) = \zero$.
  % 
  % \todo{remove}
  % \begin{remark}
  %   Having $\Yvec$ defined in Eq. \ref{eq:Y_mat} will satisfy the $\Vvec^\top \Yvec \Vvec = \zero$.
  % \end{remark}
  % 
  %   To show that, we plug in the first condition of $\Yvec$, we have
  %   $\Vvec^\top \rbr{\dvec \one^\top + \one \dvec^\top} \Vvec = \zero$.
  % % 
  % \begin{align*}
  %   \Vvec^\top \rbr{\dvec \one^\top + \one \dvec^\top} \Vvec
  %   = \rbr{\Vvec^\top \dvec \one^\top \Vvec + \Vvec^\top \one \dvec \Vvec}
  %   = \zero.
  % \end{align*}
  % % 
  %   This is simply because of $\Vvec^\top \one = \one^\top \Vvec = \zero$.

  To gain more intuition into the recipe,
  consider the barycenter problem with $\ogw_{lb}$ and only one sampled graph $\Dvec_1$.
  By Theorem \ref{thm:ogw},
  $\Bcal(\Cvec)$ can be driven to 0 and it is attained when $\hat{\Cvec}$ shares the same eigenvalues as $\hat{\Dvec}_1$,
  \ie, $\Sigma = \Delta_1$.
  Then $\hat{\Cvec}^*_{recon} = \hat{\Dvec}_1$ by our construction.
  Furthermore, the proof of Theorem \ref{thm:ogw} indicates $\one^\top \Cvec \one = \one^\top \Dvec_1 \one$
  and $\nbr{V^\top \Cvec \one} = \nbr{V^\top \Dvec_1 \one}$.
  It is then not hard to show that $\Cvec^*_{recon}$ is exactly $\Dvec_1$.
  We provide our experiments on both synthetic and real dataset in Section~\ref{subsec:exp_bary}.

  %   Due to permutation invariance,
  %   there exists a set of graphs that drive $\ogw_{lb}$ to 0.
  %   Given the optimal solution of $\Cvec^*$ for \eqref{eq:def_barycenter},
  %   represented as cost matrix from one of the permutation invariant graphs,
  %   we can retrieve the eigenvalues of its semi-projected matrix.
  %   As we can see, $\sigma_i = \delta_i, i \in \sbr{1,\cdots, m-1}$ in this scenario.
  %   Therefore, we can reconstruct the $\Cvec$ exact to be the cost matrix from sample graph,
  %   resulting the $\Bcal(\Cvec) = \ogw_{lb} (\Cvec, \Dvec)=0$ while finding the sample graph precisely.

\section{Extension to Fused $\gw$}
 \label{sec:ofgw}

 Most applications carry features for each node.
 To account for this important information,
 \citet{VayChaFlaTavetal19} proposed the fused $\gw$ ($\fgw$),
 employing an additional matrix $\Mvec$ whose $(i,k)$-th entry
 encodes the $\ell_2$ distance between the features of node $i$ in $G$ and of node $k$ in $H$.
 $\Mvec$ is asymmetric in general.
 Then the vanilla $\fgw$-distance was formulated by \citet{VayChaFlaTavetal19} as
 \begin{align}
   \label{eq:def_fgw}
    & \fgw(G,H,\Mvec)
   := \frac{\alpha}{n^2} \nbr{\Cvec}_F^2
   + \frac{\alpha}{n^2} \nbr{\Dvec}_F^2
   \\
   \nonumber
    & - \frac{1}{n^2} \max_{\Pvec \in \Ecal \cap \Ncal}
   \cbr{2 \alpha \tr(\Cvec \Pvec \Dvec \Pvec^\top)
     - (1-\alpha) \tr(\Mvec^\top \Pvec)},
 \end{align}
 where $\alpha \in \sbr{0, 1}$ is a trade-off between structure and feature measure.
 For simplicity, we will only present the treatment for two graphs of the same size.
 The extension to different sizes can be easily derived in the same way as in Section~\ref{subsec:graphs_diff_size}.
 Similar to $\Qcal$, with the additional linear term $\tr(\Mvec^\top \Pvec)$,
 there is no closed-form solution even if we replace the domain of $\Pvec$ by $\Ocal$.
 However, we can still tighten the domain from $\Ocal$ to $\Ocal \cap \Ecal$,
 % Now the new linear term $\tr(\Mvec^\top \Pvec)$ precludes the closed-form solution even if we replace the domain of $\Pvec$ by $\Ocal$.
 % Fortunately, \citet{HadRenWol92} proposed tightening the domain approximation from $\Ocal$ to $\Ocal \cap \Ecal$,
 % which although was initially intended to approximate inhomogeneous QAPs,
 % happens to be useful in our context too.
 % Compared with $\Ncal \cap \Ecal$, $\Ocal \cap \Ecal$ offers more convenience in constructing upper and lower bounds that are tight and efficient.
 leading to our new approximation
 \begin{align}
   \label{eq:obj_fgwtil}
    & \ofgw(G,H,\Mvec)
   := \frac{\alpha}{n^2} \nbr{\Cvec}_F^2
   + \frac{\alpha}{n^2} \nbr{\Dvec}_F^2
   \\
   \nonumber
    & - \frac{1}{n^2} \max_{\Pvec \in {\color{red} \Ecal \cap \Ocal}}
   \cbr{2 \alpha \tr(\Cvec \Pvec \Dvec \Pvec^\top)
     - (1-\alpha) \tr(\Mvec^\top \Pvec)}.
 \end{align}
 The pipeline of construction is illustrated in Figure~\ref{fig:transition}.
 As $\alpha$ tends to zero,
 $\ofgw$ recovers $\ogw$ between only structures.
 A number of favorable properties are enjoyed by $\ofgw$,
 which are summarized in Theorem \ref{thm:nonneg_fgw} below
 (proof deferred to Appendix~\ref{sec:proof_thm_fgwtil}).
 Although $\ogw$ must be nonnegative,
 $\ofgw$ is not guaranteed nonnegative for all $\Mvec$.
 To see a counter-example, set $\Cvec = \Dvec = \zero$ and $\Mvec = \Ivec$.
 Fortunately, for a large set of $\Mvec$, it still enjoys nonnegativity.
 \begin{theorem}
   \label{thm:nonneg_fgw}
   Suppose $M$ satisfies $\one^\top \Mvec \one \ge n ||\hat{\Mvec}||_*$.
   Then $\ofgw(G, H, \Mvec) \ge 0$ for all $G$, $H$,
   and is invariant to the (different) permutations of $G$ and $H$.
   When $\Mvec = \zero$,
   it degenerates to $\ogw$.
 \end{theorem}
 Since $\Mvec$ encodes the pairwise distance between two sets of node features $\{\Xvec_i\}_{i=1}^n$ and $\{\Yvec_i\}_{i=1}^n$
 with $\Mvec_{ij} = \nbr{\Xvec_i - \Yvec_j}^2$,
 we can confirm whether the above assumption holds a priori.
 Interestingly, this is the case in all the datasets considered in our experiment.
 For datasets without node attributes and labels, we take node degree as their features.
 In the sequel, we will make this assumption on $\Mvec$.
 %  As $\Mvec=\zero$ satisfies the assumption,
 %  $\ogw$ must always be nonnegative,
 %  a result that is not so trivial because $\Pvec \in \Ocal \cap \Ecal$ can have negative values, as opposed to $\Ecal \cap \Ncal$ in \eqref{eq:obj_qap}.

 Although $\ofgw$ is still intractable in general,
 local optimization can be performed very efficiently.
 As will be shown in Section~\ref{sec:up_low_gwtil},
 it also admits a tight {\it lower bound} using the same relaxation technique as for $\ogw$.
 Now that $\ofgw$ is motivated by computational convenience,
 one may naturally wonder whether it captures as much graph structure as the original $\fgw$ does.
 We verified this in the affirmative by following \citet{VayChaFlaTavetal19},
 where $\fgw$ is used as a kernel function served in a support vector machine.
 %  \todo{cross check the experiments}
 In Section~\ref{subsec:exp_fgw_effect},
 we will show that replacing $\fgw$ by $\ofgw$ achieves similar or better classification accuracy on a variety of datasets,
 corroborating the effectiveness of $\ofgw$.

 \subsection{Upper and lower bounds of $\ofgw$}
  \label{sec:up_low_gwtil}

  Following \eqref{eq:OE_char},
  plugging $\Pvec = \smallfrac{1}{n} \one \one^\top + \Vvec \Qvec \Vvec^\top$ into the optimization objective in \eqref{eq:obj_fgwtil} yields
  \begin{align}
    \label{eq:def_Q_fgw}
     & \max_{\Pvec \in \Ocal \cap \Ecal}
    \{2 \alpha \tr(\Cvec \Pvec \Dvec \Pvec^\top)  - (1-\alpha) \tr(\Mvec^\top \Pvec)\}
    \\
    \nonumber
     & = \smallfrac{1}{2n} s_{\Fvec} - \smallfrac{\alpha}{2n} s_{\Mvec}
    + \underbrace{\max_{\Qvec \in \Ocal}
      2 \alpha \tr (\vec\Chat \Qvec \vec\Dhat \Qvec^\top)
      + \tr (\vec\Fhat^\top \Qvec)}_{=: \ {\color{red} \Qcal(\vec\Chat, \vec\Dhat, \vec\Fhat)}},
  \end{align}
  where $\Fvec := \smallfrac{2\alpha}{n} \Cvec \one \one^\top \Dvec - \smallfrac{1-\alpha}{2\alpha} \Mvec$.
  It reveals that compared with the expression of $\gw$ in \eqref{eq:def_gw_Q},
  the additional linear term $\Mvec$ in the $\fgw$ formulation does not change the structure.
  % As a result, we can again derive the \emph{lower bound} of $\Qcal$ by using the projected quasi-Newton as before,
  % which gives rise to an \emph{upper bound} of $\ofgw$.
  As a result, we can again derive the \emph{lower bound} of $\Qcal$, \ie, \emph{upper bound} of $\ofgw$,
  by using the projected quasi-Newton as before.
  % which gives rise to an \emph{upper bound} of $\ofgw$.
  And a \emph{lower bound} of $\ofgw$ can also be obtained by decoupling the $\Qvec$ in the two terms of \eqref{eq:def_Q_fgw}.
  Specifically, by using the $\Qcal_{ub}$ defined  \eqref{eq:def_qcal_ub} via decoupling into $\Qvec_1$ and $\Qvec_2$,
  we obtain
  % 
  % \begin{align}
  %   \label{eq:def_fgwu}
  %             & \ofgw_{lb} (G, H, \Mvec) =                                         \\
  %   \nonumber & \frac{1}{n^2} \rbr{\alpha \nbr{\Cvec}_F^2 + \alpha \nbr{\Dvec}_F^2
  %     - 2 \alpha \, \Qcal_{ub}(\vec\Chat, \vec\Dhat, \vec\Fhat)
  %     -  \smallfrac{s_{\Fvec}}{n}  + \smallfrac{1-\alpha}{2\alpha n} s_{\Mvec}},
  % \end{align}
  \begin{align}
    \label{eq:def_fgwu}
     & \nonumber \ofgw_{lb} (G, H, \Mvec)
    = \frac{\alpha}{n^2} \nbr{\Cvec}_F^2
    + \frac{\alpha}{n^2} \nbr{\Dvec}_F^2
    \\
     & - \frac{1}{n^2}\sbr{2 \alpha \, \Qcal_{ub}(\vec\Chat, \vec\Dhat, \vec\Fhat)
      +  \frac{s_{\Fvec}}{n}  - \frac{1-\alpha}{2\alpha n} s_{\Mvec}}.
  \end{align}

  \begin{table*}[t]
    \centering
    \caption{Graph classification}
    \label{tab:graph_clf}
    \resizebox{1\linewidth}{!}{
      %  \begin{tabular}{ll*{11}c}
      \begin{tabular}{c|c|cc|ccc|cccc}
        \toprule
         & \multirow{2}{*}{Dataset}
         & \multicolumn{2}{c|}{Graph kernel}
         & \multicolumn{3}{c|}{GW-based SVM}
         & \multicolumn{4}{c}{OGW-based SVM}                                                                                                          \\
         & ~
         & SP                                & GK \scriptsize{($k=5$)}
         & GW                                & $\gw_{flb}$                      & FGW
         & $\ogw_{ub}$                       & $\ogw_{lb}$                      & $\ogw_{o}$                       & $\ofgw_{lb}$                     \\
        \hline
        \multirow{2}{*}{\specialcell{Vec.                                                                                                             \\Attr.}}
         & BZR
         & 78.8 \scriptsize{$\pm$ 3.3}       & 78.8 \scriptsize{$\pm$ 3.3}
         & \bf{84.9} \scriptsize{$\pm$ 1.8}  & 78.8 \scriptsize{$\pm$ 1.0}      & 84.8 \scriptsize{$\pm$ 3.2}
         & 78.8 \scriptsize{$\pm$ 1.0}       & 83.4 \scriptsize{$\pm$ 3.4}      & 83.4 \scriptsize{$\pm$ 3.4}      & 84.3 \scriptsize{$\pm$ 3.8}      \\
         & COX2
         & 78.2 \scriptsize{$\pm$ 0.4}       & 78.2 \scriptsize{$\pm$ 0.4}
         & 76.2 \scriptsize{$\pm$ 2.1}       & 78.8 \scriptsize{$\pm$ 2.2}      & 78.5 \scriptsize{$\pm$ 1.9}
         & 78.4 \scriptsize{$\pm$ 1.8}       & 78.1 \scriptsize{$\pm$ 1.8}      & 78.2 \scriptsize{$\pm$ 0.8}      & \bf{80.2} \scriptsize{$\pm$ 2.4} \\
        \hline
        \multirow{2}{*}{\specialcell{Disc.                                                                                                            \\Attr.}}
         & MUTAG
         & 78.2 \scriptsize{$\pm$ 4.1}       & 66.5 \scriptsize{$\pm$ 0.9}
         & 85.1 \scriptsize{$\pm$ 3.4}       & 60.5 \scriptsize{$\pm$ 2.3}      & \bf{85.7} \scriptsize{$\pm$ 2.4}
         & 66.5 \scriptsize{$\pm$ 2.3}       & 82.8 \scriptsize{$\pm$ 3.0}      & 82.8 \scriptsize{$\pm$ 3.0}      & 85.4 \scriptsize{$\pm$ 1.7}      \\
         & PTC-MR
         & 57.3 \scriptsize{$\pm$ 1.0}       & 55.8 \scriptsize{$\pm$ 0.7}
         & 53.4 \scriptsize{$\pm$ 4.3}       & 60.2 \scriptsize{$\pm$ 5.1}      & 51.8 \scriptsize{$\pm$ 3.4}
         & 59.5 \scriptsize{$\pm$ 10.1}      & \bf{57.9} \scriptsize{$\pm$ 4.5} & \bf{57.9} \scriptsize{$\pm$ 4.5} & 57.1 \scriptsize{$\pm$ 4.1}      \\
        \hline
        \multirow{2}{*}{\specialcell{No                                                                                                               \\Attr.}}
         & IMDB-B
         & 57.5 \scriptsize{$\pm$ 2.6}       & 60.1 \scriptsize{$\pm$ 2.4}
         & 63.4 \scriptsize{$\pm$ 0.9}       & 63.7 \scriptsize{$\pm$ 4.0}      & 65.6 \scriptsize{$\pm$ 1.8}
         & 65.1 \scriptsize{$\pm$ 0.3}       & \bf{68.3} \scriptsize{$\pm$ 1.7} & 67.4 \scriptsize{$\pm$ 1.1}      & 67.3 \scriptsize{$\pm$ 2.1}      \\
         & IMDB-M
         & 39.7 \scriptsize{$\pm$ 1.8}       & 38.2 \scriptsize{$\pm$ 2.7}
         & 47.5 \scriptsize{$\pm$ 2.3}       & 43.2 \scriptsize{$\pm$ 2.6}      & \bf{49.7} \scriptsize{$\pm$ 1.7}
         & 48.1 \scriptsize{$\pm$ 2.2}       & 48.5 \scriptsize{$\pm$ 1.9}      & 47.9 \scriptsize{$\pm$ 1.2}      & 47.1 \scriptsize{$\pm$ 2.3}      \\
        \bottomrule
      \end{tabular}
    }
  \end{table*}
  \begin{figure*}[ht]
    \vspace{-1em}
    \begin{center}
      % \hspace*{\fill}
      \subfloat[Average value of $\gw$ and its lower bounds under $\delta_g$ number of perturbations]{
        \includegraphics[width=0.32\textwidth]{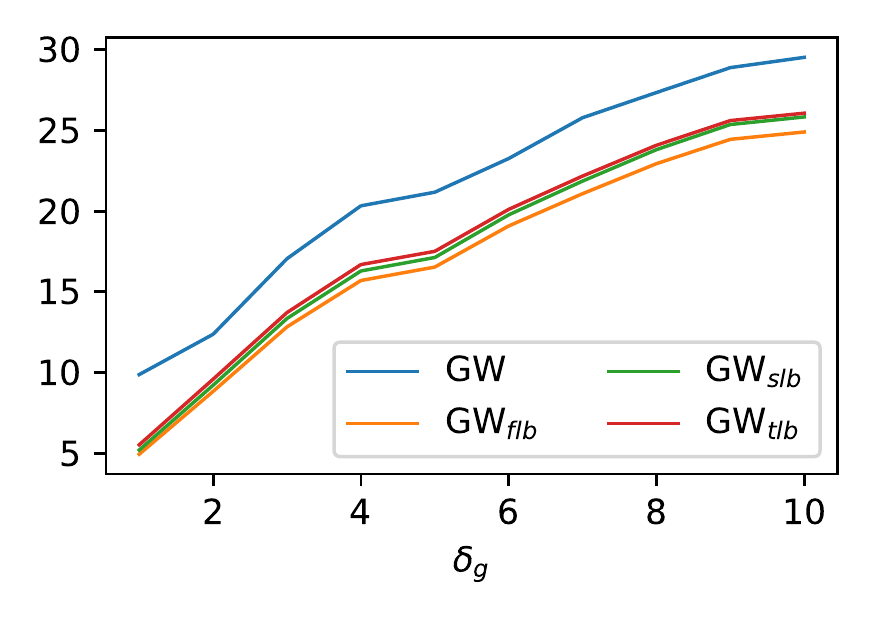}
        \label{fig:tightness_syn_gw}
      }
      ~~~
      \subfloat[Average value of $\ogw_{ub}$ and $\ogw_{lb}$ under $\delta_g$ number of perturbations]{
        \includegraphics[width=0.32\textwidth]{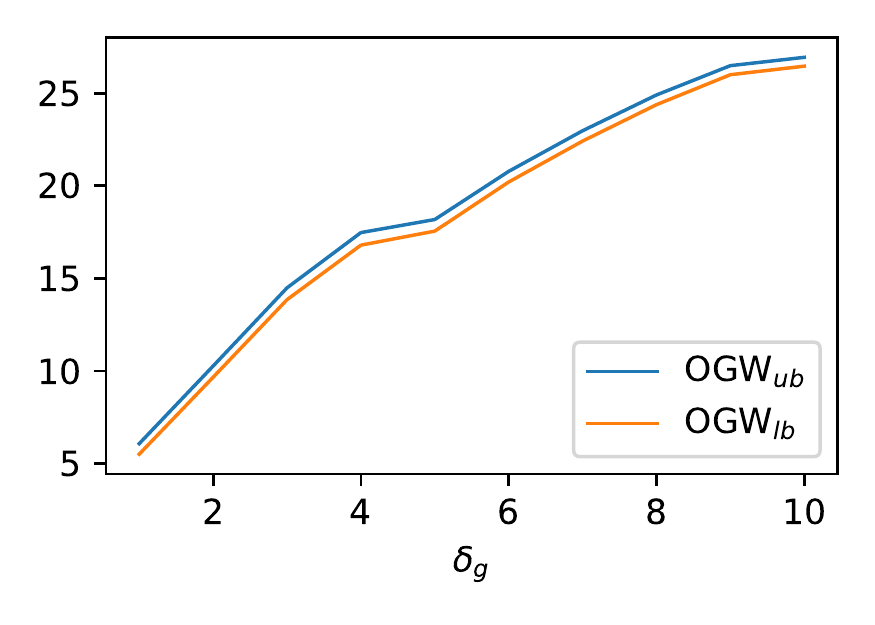}
        \label{fig:tightness_syn_ogw}
      }
      ~
      \subfloat[Running time of the lower bounds]{
        \includegraphics[width=0.32\textwidth]{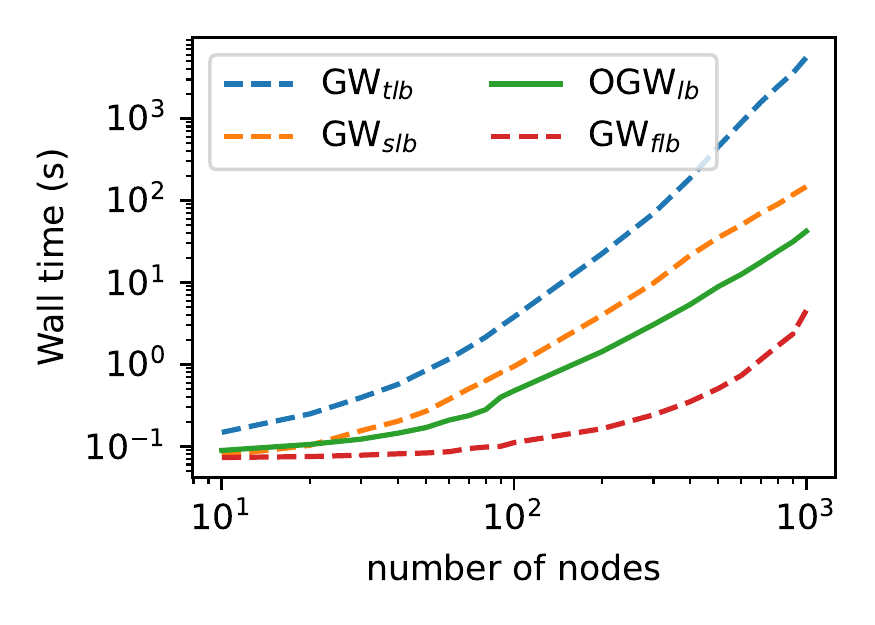}
        \label{fig:running_time}
      }
      %   \subfloat[Gap on MUTAG]{
      %     \includegraphics[width=0.32\textwidth]{images/gap_mutag.pdf}
      %     \label{fig:tightness_mutag}
      %   }
      \caption{Average value of $\gw$ and $\ogw$ as compared with their respective lower bounds to demonstrate the tightness of the lower bounds. The running time of these lower bounds is also compared.}
      % \hspace*{\fill}
      \label{fig:tightness}
    \end{center}
    \vspace{-1em}
  \end{figure*}

  \vspace{-0.9em}
\section{Experiments}
 \label{sec:experiment}

 We now demonstrate the empirical effectiveness of $\ogw$ and $\ofgw$ via two applications: graph classification and barycenter problem.
 We will also illustrate the tightness of our lower bound $\ogw_{lb}$.
 All the code and data are available at \url{https://github.com/cshjin/ogw}.

 \subsection{Effectiveness of $\ogw/\ofgw$}
  \label{subsec:exp_fgw_effect}

  Recall in Figure \ref{fig:transition},
  $\Ncal \cap \Ecal$ was replaced by $\Ocal \cap \Ecal$ because the latter enjoys tight and efficiently computable upper and lower bounds.
  So it is important to validate the resulting $\ogw / \ofgw$ as an equally good measure of comparing two graphs as the vanilla $\gw / \fgw$.
  \newpage

  \paragraph{Datasets.}
  We experimented on six graph classification datasets: BZR, COX2, MUTAG, PTC-MR, IMDB-Binary, and IMDB-Multi \citep{TUDataset}.
  % \todo{move to appendix}
  Their statistics are given in Appendix~\ref{subsec:app_dataset_stats}.
  The first four datasets contain a collection of molecules (\eg, chemical compound and ligands),
  where the vertices represent atoms and edges are chemical bonds.
  The class label represents a certain property of the molecules,
  \eg, "mutagenic effect on a specific bacterium" (MUTAG) and
  carcinogenicity of compounds for male rats (PTC-MR).
  BZR and COX2 consist collections of ligands for the benzodiazepine receptor and cyclooxygenase-2 inhibitors, respectively.
  IMDB-Binary and IMDB-Multi are the movie collaboration dataset, where nodes represent actors/actresses who played roles in movies in IMDB,
  and one edge means two played in the same movie.
  We group the dataset into three categories according to their feature property:
  vectorized, discrete, and no features.
  For the datasets with no features, we take the node degrees as their features.

  \paragraph{Settings.}
  In order to evaluate a discrepancy measure $d$,
  we follow \citet{VayChaFlaTavetal19} by studying the graph classification accuracy of an SVM,
  whose kernel $k(G, H)$ is computed by $\exp{(-\gamma d(G, H))}$.
  For both $\fgw$ and $\ofgw$,
  the feature distance matrix $\Mvec$ employed the squared Euclidean distance.
  Since $\ofgw$ itself is intractable to evaluate,
  we resort to the lower bound of $\ofgw$ in \eqref{eq:def_fgw} which has an analytic form and is nonnegative.
  For the vanilla $\gw/\fgw$, we adopt the implementation from POT package \citep{flamary2021pot},
  which initiates the transition matrix by the outer product of marginal distributions.
  And we instantiate the cost matrix $c_G$ by all-pair shortest path for each graph in the datasets,
  knowing that the structures are all connected.
  We evaluate the models by cross-validation on the hyperparameters in SVM,
  setting $\gamma$ from $\cbr{2^{-10}, 2^{-9}, \cdots, 2^{10}}$ and
  $C$ from $2^{-4}$ to $2^4$ on evenly log scale with 15 steps.
  Moreover, for the $\fgw / \ofgw$, we cross-validate the value of $\alpha$ from $\sbr{0, 1}$ with grid search.

  In addition, we consider the graph kernel methods as the baselines.
  More specifically,
  we adopt the implementation of shortest path (SP) kernel~\citep{BorKri05}
  and graphlet sampling (GK) kernel~\citep{Przulj07} from \citet{siglidis2020grakel}.
  SP kernel decomposes graphs into shortest paths and compares pairs of shortest paths according to their lengths and the labels of their endpoint,
  while GK kernel decomposes graphs into graphlets,
  \ie,\ small subgraphs with $k$ nodes where $k\in \cbr{3, 4, 5,\cdots}$,
  and counts matching graphlets in the input graphs.

  % \begin{table*}
  %   \centering
  %   \caption{Statistics of the datasets}
  %   \label{tab:datasets}
  %   \resizebox{\textwidth}{!}{%
  %     \begin{tabular}{l*{9}c}
  %       \toprule
  %       dataset & \# graphs & \# class & \# features & ave. edge & min edge & max edge & avg. node & min node & max node \\
  %       \hline
  %       BZR     & 405       & 2        & 56          & 74.0      & 26       & 120      & 35.0      & 13       & 57       \\
  %       COX2    & 467       & 2        & 38          & 86.0      & 68       & 118      & 41.0      & 32       & 56       \\
  %       MUTAG   & 188       & 2        & 7           & 38        & 20       & 66       & 17.5      & 10       & 28       \\
  %       PTC-MR  & 344       & 2        & 18          & 25.0      & 2        & 142      & 13.0      & 2        & 64       \\
  %       IMDB-B  & 1000      & 2        & -           & 130       & 52       & 2498     & 17        & 12       & 136      \\
  %       IMDB-M  & 1500      & 3        & -           & 72        & 23       & 2934     & 10        & 7        & 89       \\
  %       \bottomrule
  %     \end{tabular}}
  % \end{table*}

  \paragraph{Results.}
  The average accuracy achieved with 10-fold cross-validation is presented in Table~\ref{tab:graph_clf}.
  %   To see the effectiveness of $\ogw / \ofgw$, we compare the accuracy with the vanilla $\gw / \fgw$.
  To see the effectiveness of $\ogw$ and its lower bounds ($\ogw_{lb}$ and $\ogw_{o}$), we compare their accuracy against that of the vanilla $\gw$ and its first lower bound ($\gw_{flb}$).
  In addition, we also include $\fgw$ and $\ofgw$ that incorporate node features.
  As the table shows,
  the tractable lower bounds of our proposed metric $\ogw$ have comparable performance with $\gw$ and its variant.
  With the additional information from node features (except IMDB datasets),
  the $\ofgw$ provides higher accuracy in general.
  It corroborates that the chain in Figure~\ref{fig:transition} preserves important structures in graphs,
  %   Furthermore, it is noticeable that the accuracies with $\gw$ in the dataset of COX2 and PTC-MR are slightly worse than those in baselines.
  %   This is largely due to the intractable non-convex optimization with sub-optimal local minima in $\gw$.
  making the tractable lower bounds of $\ogw / \ofgw$ sound discrepancy measures for graphs.
  Moreover, $\ogw_{lb}$ achieves higher accuracy than $\ogw_o$ on two datasets,
  and performs similarly on the other four datasets.
  % {\color{red} To note that the $\ogw_{lb}$ and $\ogw_{o}$ have similar performance in accuracy, this is largely due to the tightness between $\ogw_{lb}$ and $\ogw_{o}$. We will provide intensive experiments in Section~\ref{subsec:tightness_lb_ub}.}

  \begin{figure*}[htbp]
    \begin{minipage}{.5\textwidth}
      \centering
      \subfloat[Gap in $\ogw$
        versus gap in $\gw$]{
        \includegraphics[width=0.53\linewidth]{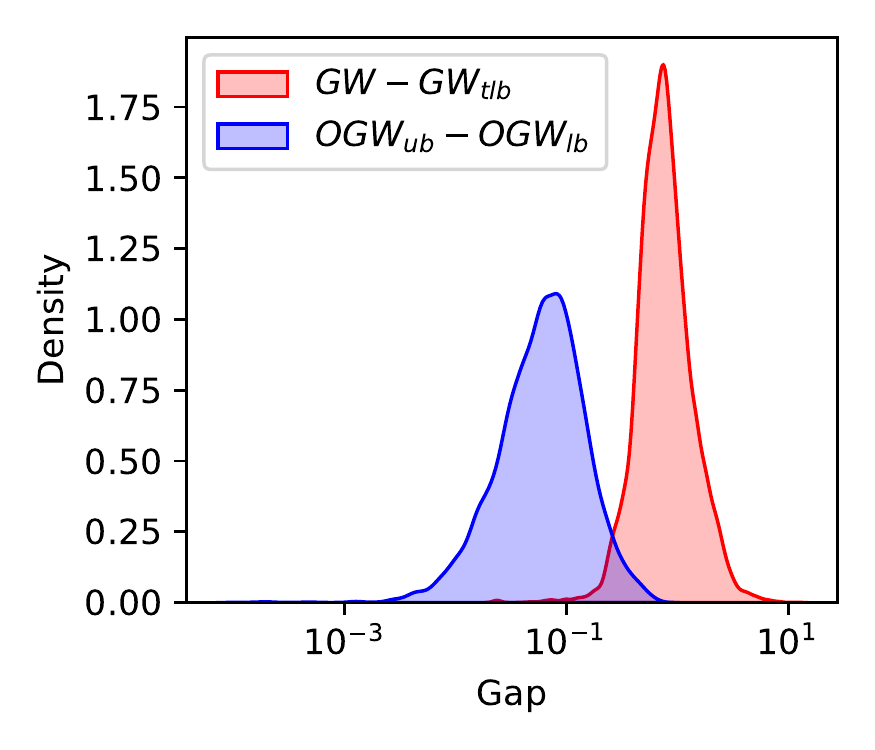}
        \label{fig:tightness_mutag}
      }
      ~
      \subfloat[Gap in $\ogw$]{
        \includegraphics[width=0.46\linewidth]{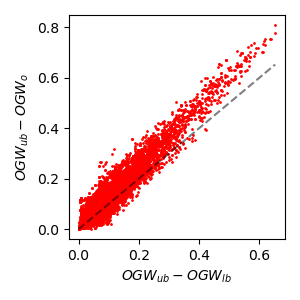}
        \label{fig:tightness_ogw_lbs_mutag}
      }
      \caption{Tightness of lower and upper bounds for $\gw$ and $\ogw$ (MUTAG dataset)}
    \end{minipage}
    ~~
    \begin{minipage}{.5\textwidth}
      \centering
      \hspace*{\fill}
      \subfloat[$\gw$]{
        \includegraphics[width=0.2\textwidth]{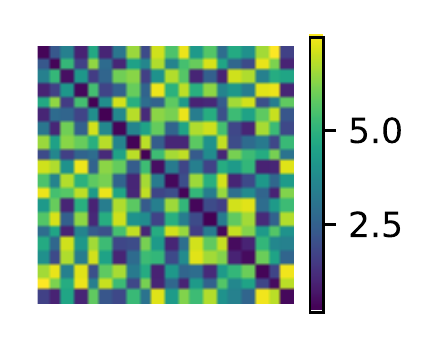}
      }
      \hfill
      \subfloat[$\ogw_{lb}$]{
        \includegraphics[width=0.2\textwidth]{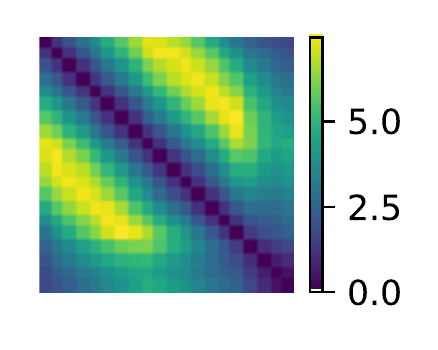}
      }
      \hfill
      \subfloat[$\ogw_{ub}$]{
        \includegraphics[width=0.2\textwidth]{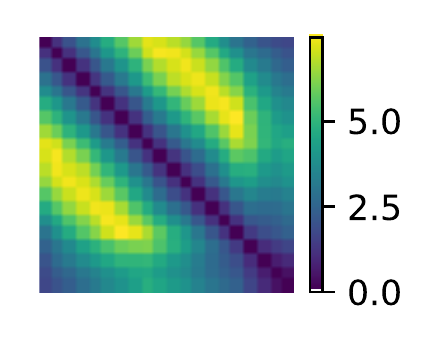}
      }
      \hspace*{\fill}
      \\
      \hspace*{\fill}
      \subfloat[$\gw$]{
        \includegraphics[width=0.16\textwidth]{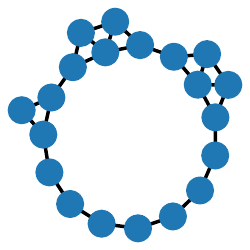}
      }
      % \qquad
      \hfill
      \subfloat[$\ogw_{lb}$]{
        \includegraphics[width=0.16\textwidth]{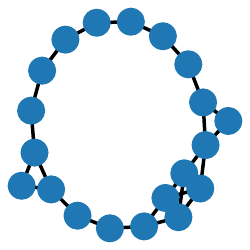}
      }
      % \qquad
      \hfill
      \subfloat[$\ogw_{ub}$]{
        \includegraphics[width=0.16\textwidth]{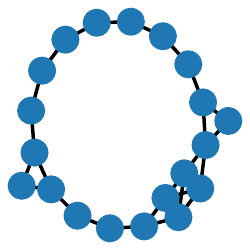}
      }
      \hspace*{\fill}
      \caption{Optimized barycenter.
        Subfigures (a-c) show the $\Cvec^*$ retrieved by optimizing the barycenter.
        Subfigures (d-f) show the reconstructed graph from the optimal cost matrix $\Cvec^*$.
      }
      \label{fig:barycenter_results_syn}
    \end{minipage}
  \end{figure*}

 \subsection{Tightness of the Lower Bound}
  \label{subsec:tightness_lb_ub}

  The tightness of our lower bound $\ogw_{lb}$, with respect to $\ogw$,
  can be evaluated through its difference to the upper bound $\ogw_{ub}$,
  which is obtained by the projected quasi-Newton method.
  Such a gap will be compared with its counterpart in $\gw$ distance,
  where the upper bound is served by a local optimizer based on the conditional gradient method,
  and the lower bounds are proposed by \citet{Memoli11},
  including FLB, SLB, and TLB (the best known lower bound).

  \paragraph{Synthetic data.}
  To demonstrate the tightness in synthetic data,
  we generate a path graph with 20 nodes and randomly perturb $\delta_g \in \sbr{1, 10}$ edges for 50 times,
  so that we can measure the distance (dissimilarity) between the original graph and the perturbed graph under different measures.
  Only connected graphs are kept.
  Figure~\ref{fig:tightness_syn_gw} and \ref{fig:tightness_syn_ogw} provide,
  respectively for $\gw$ and $\ogw$,
  the average distance as a function of the number of perturbed edges.
  The gap between $\ogw_{ub}$ and $\ogw_{lb}$ is much tighter than the best gap for the $\gw$ case, \ie, $\gw - \gw_{tlb}$.

  Note that TLB of GW requires $\Ocal(n^5)$ computational time,
  while $\ogw_{lb}$ only costs $\Ocal(n^3)$.
  To verify it, we measure the running time by varying the graph sizes.
  For each graph size that ranges from 10 to 1000, 20 Erd\H{o}s-R\'enyi random graphs are generated, ensuring they are connected.
  Then their average running time is reported in Figure~\ref{fig:running_time}, which clearly matches the analyzed computational complexity.

  % \todo{rephrase}
  \paragraph{Real-world data.}
  We also examine the gap between lower and upper bounds on the real-world dataset MUTAG by evaluating pairwise distance.
  % We evaluate the pairwise distance between graphs with different methods.
  % For $\ogw$, we measure the gap between its lower and upper bound as before.
  % For $\gw$, we evaluate the 
  % We evaluate the pairwise distance between graphs under $\gw$ and $\gw_{tlb}$ (best known lower bound).
  % \todo{rephrase}
  % And measure the gap between $\ogw_{ub}$ and $\ogw_{lb}$ for $\ogw$ the same as before.
  Figure~\ref{fig:tightness_mutag} provides the distribution of the gaps from $\gw$ and $\ogw$.
  For $\gw$, we report the gap between $\gw$ local optimizer and $\gw_{tlb}$.
  % To point out, we omit the gap less than $1e^{-3}$.
  Clearly, the gap between $\ogw_{ub}$ and $\ogw_{lb}$ centers around $0.1$,
  while that between the $\gw$ and its TLB concentrates around $1$.

  In addition, we also compare in Figure~\ref{fig:tightness_ogw_lbs_mutag} the tightness of $\ogw_{lb}$ and $\ogw_{o}$,
  both as a lower bound for $\ogw$.
  Each point in the scatter plot represents a graph in the MUTAG dataset, and the horizontal (resp. vertical) axis is the gap between the upper bound of $\ogw$ and $\ogw_{lb}$ (resp. $\ogw_o$).
  The fact that the vast majority of the points lie above the diagonal confirms the superior tightness achieved by $\ogw_{lb}$.
  This demonstrates the benefit of employing $\Ecal$ in the constraint and separating $\Qvec_1$ and $\Qvec_2$ in \eqref{eq:def_qcal_ub}.
  %   It also verifies the tightness of $\ogw_{lb}$ and $\ogw_{o}$ in the context of classification from Table~\ref{tab:graph_clf}. 
  More tightness results on other datasets are provided in Appendix~\ref{subsec:app_tightness}.

  \vspace{-2em}
 \subsection{Barycenter}
  \label{subsec:exp_bary}

  In the last set of experiments, we evaluate the ability of $\ogw_{lb}$ and $\ogw_{ub}$ to solve the barycenter problem.

  % \todo{rephrase}
  \vspace{-0.2em}
  \paragraph{Synthetic data.}
  To start with, we generate a set of cycle-like graphs with different sizes ranging from 15 to 25.
  In addition, we also explicitly add random structural noise to the graphs separately,
  and pre-compute their shortest path as cost matrices.
  All synthetic samples are plotted in Appendix~\ref{subsec:app_bary_samples}.
  To initialize the barycenter, we fix the number of nodes to be 20 and
  initiate a random symmetric $\Cvec$ when starting the block coordinate descend to update $\Pvec$.
  For $\gw$, we adopt the implementation from \citet{PeyCutSol16} to find the optimal cost matrix of the center.
  For the $\ogw_{lb}$ and $\ogw_{ub}$, we solve the problem by our proposed eigen projection method in Section~\ref{subsec:barycenter}.

  To better visualize the results,
  we also reconstruct the adjacency matrix following a standard heuristic \citep{VayChaFlaTavetal19}.
  In particular, for a given threshold,
  a pair of nodes can be connected by an edge if and only if their corresponding entry in the cost matrix is below the threshold.
  Then we perform a line search on the threshold to minimize the difference between the optimal cost matrix and the one corresponding to the threshold based adjacency matrix.
  From Figure~\ref{fig:barycenter_results_syn},
  we can see the reconstructed cost matrix for the barycenter from our $\ogw_{lb}$ and $\ogw_{ub}$ are well aligned to its node ordering.
  We also recognize that the result from $\gw$ is just one of the local minimizers in the context of permutation.
  Due to the tightness between the lower and upper bounds,
  it is also hard to differentiate the structures found by $\ogw_{lb}$ and $\ogw_{ub}$ (i.e., sub-figures e and f).

  % REVIEW: use table?
  \begin{table}[t]
    \centering
    \caption{Barycenters from point cloud MNIST-2D samples}
    \label{tab:barycenter_results_mnist}
    % \resizebox{0.45\textwidth}{!}{
    \begin{tabular}{ c|ccc}
                            & $\gw$
                            & $\ogw_{lb}$
                            & $\ogw_{ub}$                                                                  \\
      \hline
      \rotatebox{90}{$y=0$} & \includegraphics[width=0.06\textwidth]{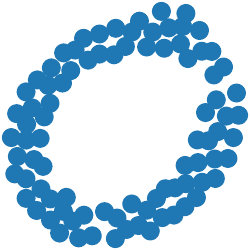}
                            & \includegraphics[width=0.06\textwidth]{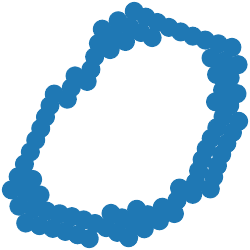}
                            & \includegraphics[width=0.06\textwidth]{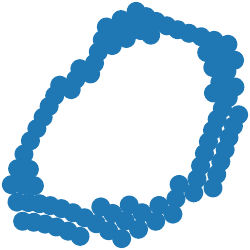} \\
      \rotatebox{90}{$y=6$} & \includegraphics[width=0.06\textwidth]{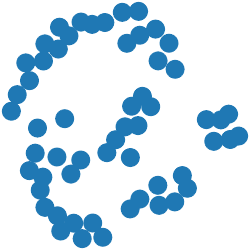}
                            & \includegraphics[width=0.06\textwidth]{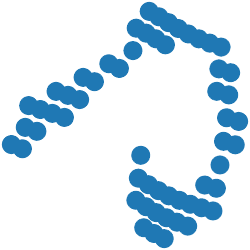}
                            & \includegraphics[width=0.06\textwidth]{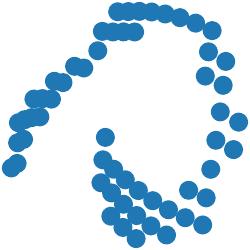} \\
      \rotatebox{90}{$y=9$} & \includegraphics[width=0.06\textwidth]{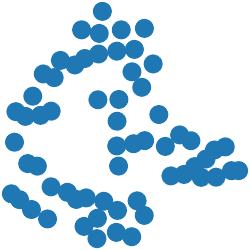}
                            & \includegraphics[width=0.06\textwidth]{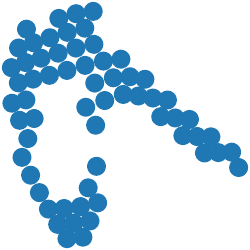}
                            & \includegraphics[width=0.06\textwidth]{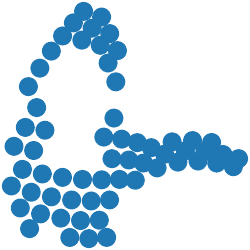} \\
    \end{tabular}
    % }
  \end{table}

  \vspace{-0.2em}
  \paragraph{Point cloud data.}
  We also explore the barycenter on point cloud dataset called MNIST-2D
  \footnote{Point cloud MNIST-2D dataset: \url{https://www.kaggle.com/cristiangarcia/pointcloudmnist2d}}.
  Refer to Appendix~\ref{subsec:app_bary_samples} for the plots of the corresponding point cloud data.
  Clearly, the point clouds for 6 and 9 are quite similar up to rotation.

  The point cloud is modeled by a graph whose nodes correspond to non-zero (non-black) pixels,
  represented by their 2D coordinates.
  Without constructing the explicit mesh connections between pixels,
  we take the Euclidean distance as their cost matrix $\Dvec_i$.
  After retrieving the optimal $\Cvec^*$ of the barycenter,
  we further uncover the associated optimal coordinates.
  A BFGS optimizer was used to seek the locally optimal coordinates such that the resulting Euclidean distance between pixels best reconstructs $\Cvec^*$.
  Clearly, such a recovery can only be up to the standard invariant transformations such as rotation and shift,
  and they are determined by the random initialization of BFGS.

  Table~\ref{tab:barycenter_results_mnist} illustrates the optimal point clouds with labels to be $0$, $6$, and $9$.
  We sample 5 different point clouds for each digit and set the weight $\lambda_i$ to the uniform distribution.
  Moreover, we fix the number of nodes on the barycenter as the minimum size from the samples.
  Compared with $\gw$,
  $\ogw_{lb}$ and $\ogw_{ub}$ clearly find better point clouds of the barycenter in the cases of digit 6 and 9.
  Moreover, thanks to the tight gap between the lower bounds,
  the performance of $\ogw_{lb}$ and $\ogw_{ub}$ differ only indistinguishably.

  % \begin{contributions} % will be removed in pdf for initial submission,
  %   % so you can already fill it to test with the
  %   % ‘accepted’ class option
  %   Briefly list author contributions.
  %   This is a nice way of making clear who did what and to give proper credit.

  %   H.~Q.~Bovik conceived the idea and wrote the paper.
  %   Coauthor One created the code.
  %   Coauthor Two created the figures.
  % \end{contributions}

  \vspace{-0.5em}
  \begin{acknowledgements} % will be removed in pdf for initial submission,
    % so you can already fill it to test with the
    % ‘accepted’ class option
    % Briefly acknowledge people and organizations here.

    % \emph{All} acknowledgements go in this section.

    We thank the reviewers for their constructive comments.
    This work is supported by NSF grant RI:1910146.
  \end{acknowledgements}

  \bibliography{bibfile}

  \onecolumn
  \appendix

\section{Proofs and detailed algorithms}

 \subsection{Value of $\mathbf{V}$}
  \label{sec:Vmatrix}

  As proposed by \citet{HadRenWol92},
  let $x = -\smallfrac{1}{n+\sqrt{n}}$ and
  $y = -\smallfrac{1}{\sqrt{n}}$.
  Then $\Vvec \in \RR^{n \times (n-1)}$ can be assigned as follows.
  Set the first row to $y \one_{n-1}^\top$,
  and the remaining $n-1$ rows, as an $(n-1) \times (n-1)$ matrix, to $x \one_{n-1} \one_{n-1}^\top+ \Ivec_{n-1}$.

 \subsection{Proof of Theorem \ref{thm:ogw}}
  \label{sec:proof_nonneg_ogw}
  First, we prove the non-negativity of $\ogw$ and $\ogw_{lb}$,
  as that for $\ogw_o$ is trivial from \eqref{eq:obj_ogw_o}.
  Since $\ogw_{lb}$ lower bounds $\ogw$, it suffices to prove the non-negativity of the former.
  Denote $\Evec = \Cvec \one \one^\top \Dvec$.
  Then
  \begin{align}
    \ogw_{lb} (\Cvec, \Dvec)
    = \frac{1}{n^2} \norm{\Cvec}_F^2
    + \frac{1}{n^2} \norm{\Dvec}_F^2
    - \frac{2}{n^2} \sbr{\max_{\Qvec_1} \tr (\hat{\Cvec} \Qvec_1 \hat{\Dvec} \Qvec_1^\top)
      + \frac{2}{n}\max_{\Qvec_2}\tr (\hat{\Evec}^\top \Qvec_2)
      + \frac{1}{n^2} s_{\Cvec} s_{\Dvec}}.
  \end{align}
  Let $\uvec = \one / \sqrt{n} \in \RR^{n}$. Then $\Vvec^\top \uvec = \zero$ and
  \begin{align}
    \nbr{\Cvec}^2_F
    = \nbr{
      \begin{pmatrix}
        \Vvec^\top \\
        \uvec^\top
      \end{pmatrix}
      \Cvec
      \begin{pmatrix}
        \Vvec, \uvec
      \end{pmatrix}
    }_F^2
    = \nbr{
      \begin{pmatrix}
        \Vvec^\top \Cvec \Vvec & \Vvec^\top \Cvec \uvec \\
        \uvec^\top \Cvec \Vvec & \uvec^\top \Cvec \uvec
      \end{pmatrix}
    }_F^2
    = \nbr{\hat{\Cvec}}^2_F + 2 \nbr{\Vvec^\top \Cvec \uvec}^2 + (\uvec^\top \Cvec \uvec)^2.
  \end{align}

  It is easy to see that
  \begin{align}
    \max_{\Qvec_1 \in \Ocal} \tr (\hat{\Cvec} \Qvec_1 \hat{\Cvec} \Qvec_1^\top) = \nbr{\hat{\Cvec}}_F^2,
    \qquad
    \max_{\Qvec_2 \in \Ocal} \tr (\hat{\Evec} \Qvec_2^\top) =
    \nbr{\Vvec^\top \Cvec \uvec} \nbr{\Vvec^\top \Dvec \uvec}.
  \end{align}

  Denoting $\lambda_i(\hat{\Cvec})$ as the $i$-th largest eigenvalue of $\hat{\Cvec}$, we have
  \begin{align}
    \ogw_{lb} (\Cvec, \Dvec)
    = & \frac{1}{n^2} \rbr{
      \nbr{\hat{\Cvec}}^2_F + 2 \nbr{\Vvec^\top \Cvec \uvec}^2 + (\uvec^\top \Cvec \uvec)^2
      + \nbr{\hat{\Dvec}}^2_F + 2 \nbr{\Vvec^\top \Dvec \uvec}^2 + (\uvec^\top \Dvec \uvec)^2
    }                        \\
      & -\frac{1}{n^2} \rbr{
      2 \sum_i \lambda_i(\hat{\Cvec}) \lambda_i(\hat{\Dvec})
      + \frac{4}{n} \nbr{\Vvec^\top \Cvec \one} \nbr{\Vvec^\top \Dvec \one}
      + \frac{2}{n^2} (\one^\top \Cvec \one) (\one^\top \Dvec \one)
    }.
  \end{align}
  The non-negativity can then be easily derived by summing up the following equalities,
  which are all implied by $a^2 + b^2 \ge 2 ab$:
  \begin{align}
    \label{eq:ogw_lb_ineq1}
    \nbr{\hat{\Cvec}}^2_F + \nbr{\hat{\Dvec}}^2_F
    = \sum_i \lambda_i(\hat{\Cvec})^2 + \sum_i \lambda_i(\hat{\Dvec})^2
    \  & \ge \  2 \sum_i \lambda_i(\hat{\Cvec}) \lambda_i(\hat{\Dvec}),             \\
    \label{eq:ogw_lb_ineq2}
    \nbr{\Vvec^\top \Cvec \uvec}^2_F + \nbr{\Vvec^\top \Dvec \uvec}^2_F
    \  & \ge \ \frac{2}{n} \nbr{\Vvec^\top \Cvec \one} \nbr{\Vvec^\top \Dvec \one}, \\
    \label{eq:ogw_lb_ineq3}
    (\uvec^\top \Cvec \uvec)^2 + (\uvec^\top \Dvec \uvec)^2 \
       & \ge \ \frac{2}{n^2} (\one^\top \Cvec \one) (\one^\top \Dvec \one).
  \end{align}
  % 
  % \todo{verify claim}

  Next, let us assume the two graphs are isomorphic, \ie,
  there exists a permutation matrix $\vec{\Pi}$
  such that $\Cvec = \vec{\Pi}^\top \Dvec \vec{\Pi}$.
  %   {\color{red}It is easy to see that the equality holds with the property
  %   $\lambda_i(\Vvec^\top \Cvec \Vvec) = \lambda_i(\Vvec^\top \vec{\Pi}^\top \Dvec \vec{\Pi} \Vvec) = \lambda_i(\Vvec^\top \Dvec \Vvec)$,
  %   \ie, two graphs share the same eigen systems.
  %   Therefore, the $\ogw_{lb}(\Cvec, \Dvec) = 0$ if two graphs are isomorphic.}
  %
  For $\ogw$, simply pick $\Pvec = \vec{\Pi}^\top$, which is in $\Ocal \cap \Ecal$.
  Then $n^2$ times of the right-hand side of \eqref{eq:obj_gwtil} equals
  \begin{align}
    \nbr{\vec{\Pi}^\top \Dvec \vec{\Pi}}^2_F + \nbr{\Dvec}_F^2 - 2 \tr(\vec{\Pi}^\top \Dvec \vec{\Pi} \vec{\Pi}^\top \Dvec \vec{\Pi})
    = \nbr{\Dvec}^2_F + \nbr{\Dvec}_F^2 - 2 \tr(\Dvec \Dvec) = 0.
  \end{align}
  Since $\ogw \ge 0$, $0$ must be its value in this case.
  As $\ogw_{lb}$ and $\ogw_o$ lower bound $\ogw$ and they are nonnegative,
  they must also be 0.

  Finally, we will show the triangle inequality.
  For $\ogw$, let graphs $G$, $H$, $J$ share the same size and have distance matrices $\Cvec, \Dvec, \Evec$, respectively.  Denote $d(\Cvec,\Dvec) = \sqrt{\ogw(G,H)}$ and its corresponding $\Pvec$ in \eqref{eq:obj_gwtil} is $\Pvec_1$.  Likewise, denote $d(\Dvec,\Evec) = \sqrt{\ogw(H,J)}$ with $\Pvec_2$. Then $d(\Cvec,\Dvec) = \nbr{\Pvec_1^\top \Cvec \Pvec_1 - \Dvec}_F$ and $d(\Dvec,\Evec) = \nbr{\Pvec_2 \Evec \Pvec_2^\top - \Dvec}_F$. Now let $\Pvec_3 = \Pvec_1 \Pvec_2$, which is obviously in $\mathcal{O} \cap \mathcal{E}$. Finally,
  \begin{align}
    d(\Cvec,\Evec) & \le \nbr{\Pvec_3^\top \Cvec \Pvec_3 - \Evec}_F = \nbr{\Pvec_1^\top \Cvec \Pvec_1 - \Pvec_2 \Evec \Pvec_2^\top}_F               \\
                   & \le \nbr{\Pvec_1^\top \Cvec \Pvec_1 - \Dvec}_F + \nbr{\Pvec_2 \Evec \Pvec_2^\top - \Dvec}_F = d(\Cvec,\Dvec) + d(\Dvec,\Evec).
  \end{align}

  For $\ogw_{lb}$, \eqref{eq:ogw_lb_ineq1} to \eqref{eq:ogw_lb_ineq3} effectively decompose the values of $\ogw_{lb}$ as the sum of ``left-hand side minus right-hand side'' from the three equations. Therefore, $\ogw_{lb}$ equals the squared $\ell_2$ distance between the feature representations of the two graphs, where the first graph (and likewise the second graph) is represented as $(\lambda_1(\hat{\Cvec}), \ldots, \lambda_{n-1}(\hat{\Cvec}), \nbr{\Vvec^\top \Cvec \one} / \sqrt{n}, \one^\top \Cvec \one/n)^\top$. This directly leads to the triangle inequality over graphs.

  $\ogw_o$ given in \eqref{eq:obj_ogw_o} obviously satisfies the triangle inequality,
  because it represents each graph with its sorted eigenvalues, and computes their Euclidean distance.

  % \todo{Verify: necessary but not sufficient condition: $\ogw_{lb} = 0$  cannot imply the isomorphism.}
  % Now, assume $\ogw_{lb}(\Cvec, \Dvec) = 0$, where $\Cvec, \Dvec$ are the distance matrices from two graphs.
  % We can see that $\hat{\Cvec}, \hat{\Dvec}$ are similar (sharing the same eigen values),
  % \ie, there exists an invertible matrix $\Pvec$, such that $\hat{\Cvec} = \Pvec^{-1} \hat{\Dvec} \Pvec$.
  % \todo{incomplete}

  % REVIEW: https://math.stackexchange.com/questions/1520807/not-isomorphic-graphs-with-same-spectrum-exists?rq=1

 \subsection{Proof of Theorem \ref{thm:nonneg_fgw}}
  \label{sec:proof_thm_fgwtil}
  Following the idea of Theorem~\ref{thm:ogw}, we prove the non-negativity of $\ofgw$ as follows:
  \begin{align}
           & \ofgw_{lb}(G,H, \Mvec) \\
    = \    & \frac{1}{n^2} \sbr{
      \alpha \nbr{\Cvec}_F^2 + \alpha \nbr{\Dvec}_F^2
      - 2 \alpha \max_{\Qvec_1 \in \Ocal} \tr(\vec\Chat \Qvec_1 \vec\Dhat \Qvec_1^\top)
      - 2 \nbr{\alpha \vec\Ehat - \frac{1-\alpha}{2\alpha} \vec\Mhat}_*
      -  \frac{\alpha}{n} s_{\Evec}
      + \frac{1-\alpha}{\alpha n} s_{\Mvec}
    }                               \\
    \ge \  & \frac{1}{n^2} \sbr{
      \alpha \nbr{\Cvec}_F^2 + \alpha \nbr{\Dvec}_F^2
      - 2 \alpha \max_{\Qvec_1 \in \Ocal} \tr(\vec\Chat \Qvec_1 \vec\Dhat \Qvec_1^\top)
      - 2 \alpha \nbr{\vec\Ehat}_*
      -  \frac{\alpha}{n} s_{\Evec}
      - \frac{1-\alpha}{\alpha} \nbr{\vec\Mhat}_*
      + \frac{1-\alpha}{\alpha n} s_{\Mvec}
    }                               \\
    \ge \  & \frac{1}{n^2} \sbr{
      - \frac{1-\alpha}{\alpha} \nbr{\vec\Mhat}_*
      + \frac{1-\alpha}{\alpha n} s_{\Mvec}
    }                               \\
    \ge \  & 0.
  \end{align}
  Here the first $\ge$ is by the triangle inequality of the trace norm,
  the second $\ge$ is by $\ogw_{lb}(G,H) \ge 0$ in Theorem~\ref{thm:ogw},
  and the third is by the assumption on $\Mvec$ in Theorem~\ref{thm:nonneg_fgw},
  \ie, $\nbr{\Mvec}_* \ge \one^\top \Mvec \one$.

  When $\Mvec = \zero$, $\ofgw(G, H, \Mvec) = 0$ because the optimal $\Pvec$ in the definition of Eq.~\eqref{eq:obj_fgwtil} is identity when $\Cvec = \Dvec$.

 \subsection{Proof of Theorem \ref{thm:nonsquare_reparam} regarding the parametrization of non-square matrices}
  \label{subsec:app_p_param_nonsquare}

  \textbf{Sufficiency}.  One can easily check that
  \begin{align}
    \Pvec^\top \Pvec = \smallfrac{1}{n} \one_n \one_n^\top + \Vvec \Vvec^\top = \Ivec_{n},
  \end{align}
  where the last equality is by
  $\Vvec^\top \one = \zero$ and $\Vvec^\top \Vvec = \Ivec_{n-1}$.
  \\

  \noindent
  \textbf{Necessity}.
  By \citet[Lemma 3.1]{HadRenWol92}, we can rewrite $\Pvec$ as
  \begin{align}
    \Pvec & = \Bvec_m
    \begin{bmatrix}
      1     & \zero \\
      \zero & \Qvec
    \end{bmatrix}
    \Bvec_n^\top,
  \end{align}
  where $\Bvec_m =
    \begin{bmatrix}
      \frac{\one_m}{\sqrt{m}} & \Uvec
    \end{bmatrix}$,
  $\Bvec_n =
    \begin{bmatrix}
      \frac{\one_n}{\sqrt{n}} & \Vvec
    \end{bmatrix}$.
  As a result,
  \begin{align}
    \Pvec^\top \Pvec & =
    \begin{bmatrix}
      \one & \Vvec
    \end{bmatrix}
    \begin{bmatrix}
      1     & \zero^\top       \\
      \zero & \Qvec^\top \Qvec
    \end{bmatrix}
    \begin{bmatrix}
      \one^\top \\  \Vvec^\top
    \end{bmatrix} = \Ivec_n.
  \end{align}

 \subsection{From uniform distribution to non-uniform distribution}
  \label{subsec:app_node_dist}
  In the setup of $\ogw$,
  we set, for simplicity, the node distribution as uniform, \ie,
  \begin{align}
    \tilde{\Ocal} \cap \tilde{\Ecal}
    := \cbr{\Pvec^\top \Pvec = \Ivec_n,
      \Pvec\one_n = \sqrt{\frac{n}{m}}\one_m,
      \Pvec^\top \one_m = \sqrt{\frac{n}{m}}\one_m,
      \Pvec \in \RR^{m \times n}}.
  \end{align}
  To consider the non-uniform distribution,
  we can simply replace the uniform distribution with any arbitrary distribution $\pvec$ and $\qvec$ by
  \begin{align}
    \tilde{\Ocal} \cap \tilde{\Ecal}
    := \cbr{\Tvec^\top \Tvec = \frac{\Ivec_n}{mn}, \Tvec\one_n = \pvec, \Tvec^\top \one_m = \qvec, \pvec \in \RR^{m}, \qvec \in \RR^{n}}.
  \end{align}
  Note that the semi-orthogonal domain is replaced by a scalar, without hurting the tractable algorithm to provide bounded results.

 \subsection{From $\ell_2$ loss to KL divergence}
  \label{subsec:app_kl_divergence}
  Let us recall the definition $\gw$ in Eq. \eqref{eq:def_gw}, replacing $\ell_2$ loss with KL divergence
  $KL(a | b) = a \log (a/b) - a + b$
  will lead the problem to be rewritten in the Koopmans-Beckmann form
  \begin{align}
    \gw_{KL}(\Cvec, \Dvec)
    = \frac{\tr(\Cvec \log(\Cvec) - \Cvec)}{m^2}
    + \frac{\tr(\Dvec)}{n^2}
    - \max_{\Tvec \in \Ecal \cap \Ncal} \tr(\Cvec \Tvec \log(\Dvec) \Tvec^\top).
  \end{align}
  Note that we stick with the original domain of $\Ecal$, \ie, $\Tvec\one = \pvec, \Tvec^\top \one = \qvec$,
  where there is no constant scalar in front of the last term.
  Therefore, consider the partial derivative of $\gw$ w.r.t. $\Cvec$ is presented as
  \begin{align}
    \frac{\partial \gw_{KL}}{\partial \Cvec}
      = \frac{\log \Cvec}{m^2}
      - \Tvec^\top \log(\Dvec) \Tvec
      = \zero
      \implies \Cvec^* = e^{m^2 \Tvec^\top \log(\Dvec) \Tvec},
  \end{align}
  where the optimal $\Cvec^*$ is used in the block coordinate relaxation when solving the barycenter problem.
  % \todo{numerical stability in POT, 0 in C, D}
  To consider the
  \begin{align}
    \ogw_{KL} (\Cvec, \Dvec)
    =\frac{\tr(\Cvec \log(\Cvec) - \Cvec)}{m^2}
    + \frac{\tr(\Dvec)}{n^2}
    - \frac{1}{mn} \max_{\Pvec \in \Ecal \cap \Ocal} \tr(\Cvec \Pvec \log(\Dvec) \Pvec^\top),
  \end{align}
  where $\log(\Cvec), \log(\Dvec)$ are evaluated as the inputs,
  we can still apply the analytical methods in Section~\ref{subsec:ub_lb_ogw} to achieve $\ogw_{lb}$ and $\ogw_{ub}$.

  For tractable lower bound / upper bound,
  Let $\Pvec = \frac{1}{\sqrt{mn}} \one \one^\top + \Uvec \Qvec \Vvec^\top$, we have
  \begin{align}
    \max_{\Pvec \in \Ocal \cap \Ecal} \tr(\Cvec \Pvec \log(\Dvec) \Pvec^\top)
    = \max_{\Qvec \in \widetilde{\Ocal}} \tr\sbr{
      \Cvec \rbr{\frac{1}{\sqrt{mn}} \one \one^\top + \Uvec \Qvec \Vvec^\top}
      \log \Dvec \rbr{\frac{1}{\sqrt{mn}} \one \one^\top + \Uvec \Qvec \Vvec^\top}^\top}.
  \end{align}
  This is exactly the same as the $\ell_2$ loss with the replacement of $\Dvec$ with $\log (\Dvec)$.
  Therefore, the same lower bound and upper bound can be applied to the loss function with KL divergence.

  For the barycenter problem, we can take partial derivative of $\ogw$ w.r.t. $\Cvec$,
  \begin{align}
    \frac{\partial \ogw_{KL}}{\partial \Cvec}
      = \frac{\log \Cvec}{m^2}
      - \frac{1}{mn}\Pvec^\top \log(\Dvec) \Pvec
      = \zero
      \implies \Cvec^* = e^{\smallfrac{m}{n} \Pvec^\top \log(\Dvec) \Pvec}.
  \end{align}
  Consider the $\ogw_{ub}$, we recover the $\Pvec$ from the optimal $\Qvec$ from $\Qcal_{lb}$ by joint optimize the quadratic and linear term.
  And for the $\ogw_{lb}$, consider
  \begin{align}
    \Qcal_{ub}
     & = \max_{\Qvec_1 \in \Ocal} \tr(\vec\Chat \Qvec_1 \vec\Dhat \Qvec_1^\top)
    + \max_{\Qvec_2 \in \Ocal} \tr(\vec\Ehat^\top \Qvec_2)                                      \\
     & = \max_{\Qvec_1} \tr(\Vvec^\top \Cvec \Vvec \Qvec_1 \Vvec^\top \Dvec \Vvec \Qvec_1^\top)
    +\max_{\Qvec_2} \tr(\smallfrac{2}{n}\Dvec^\top\one\one^\top \Cvec \Qvec_2),
  \end{align}
  the partial derivative of $\ogw_{lb}$ w.r.t. $\Cvec$ goes to
  \begin{align}
     & \frac{\partial \ogw_{KL, lb}}{\partial \Cvec}
      = \frac{\log \Cvec}{m^2}
    \\
      \nonumber
     & - \frac{2}{mn} \rbr{
        \frac{\one_m \one_m^\top s_{\log \Dvec}}{mn}
        + \frac{1}{\sqrt{mn}} \rbr{\one_m \one_n^\top \log \Dvec \Vvec \Qvec_2^\top \Uvec^\top
          + \Uvec \Qvec_2 \Vvec^\top \log \Dvec \one_n \one_m^\top}
        + \Uvec \Qvec_1 \Vvec^\top \log \Dvec \Vvec \Qvec_1^\top \Uvec^\top
      },
  \end{align}
  which serves the closed-form solution of $\Cvec$ when applying the block coordinate update in barycenter.

 \subsection{Derivative of Barycenter w.r.t. Cost Matrix}
  \label{sec:grad_barycenter_ogw_lb}
  Consider the quadratic term in \eqref{eq:obj_gwtil}, let's take the partial derivative respect to $\Cvec$,
  \begin{align}
    \frac{\partial }{\partial \Cvec} \tr (\Cvec \Pvec \Dvec \Pvec^\top) = \Pvec \Dvec \Pvec^\top.
  \end{align}
  Noting that the $\ogw_{ub}$ solves the $\Qcal$ jointly by projected quasi-Newton method.
  Having
  \begin{align}
    \Qvec^* \leftarrow \max_{\Qvec \in \Ocal} \tr(\vec\Chat \Qvec \vec\Dhat \Qvec^\top) + \tr(\vec\Ehat^\top \Qvec),
  \end{align}
  we can retrieve the gradient information directly by replacing $\Pvec = \smallfrac{1}{\sqrt{mn}} \one_m \one_n^\top + \Uvec \Qvec^* \Vvec^\top$.

  For the $\ogw_{lb}$,
  by plug into the $\Pvec \Dvec \Pvec^\top$, we have
  \begin{align}
    \Pvec \Dvec \Pvec^\top & = (\smallfrac{1}{\sqrt{mn}} \one_m \one_n^\top + \Uvec \Qvec \Vvec^\top)
    \Dvec
    (\smallfrac{1}{\sqrt{mn}} \one_n \one_m^\top + \Vvec \Qvec^\top \Uvec^\top)
    \\
                           & = \smallfrac{1}{mn} \one_m \one_n^\top \Dvec \one_n \one_m^\top
    + \smallfrac{1}{\sqrt{mn}} \one_m \one_n^\top \Dvec \Vvec \Qvec^\top \Uvec^\top
    + \smallfrac{1}{\sqrt{mn}} \Uvec \Qvec \Vvec^\top \Dvec  \one_n \one_m^\top
    + \Uvec \Qvec \Vvec^\top \Dvec \Vvec \Qvec^\top \Uvec^\top
    \\
                           & = \smallfrac{s_{\Dvec}}{mn} \one_m \one_m^\top
    + \smallfrac{1}{\sqrt{mn}}\rbr{\one_m \one_n^\top \Dvec \Vvec \Qvec^\top \Uvec^\top
      + \Uvec \Qvec \Vvec^\top \Dvec  \one_n \one_m^\top}
    + \Uvec \Qvec \Vvec^\top \Dvec \Vvec \Qvec^\top \Uvec^\top.
  \end{align}
  And solving the decoupled two terms will have
  \begin{align}
    \Qvec_1^*, \Qvec_2^* \leftarrow \max_{\Qvec_1 \in \Ocal} \tr(\vec\Chat \Qvec_1 \vec\Dhat \Qvec_1^\top) + \max_{\Qvec_2 \in \Ocal} \tr(\vec\Ehat^\top \Qvec_2),
  \end{align}
  to update the partial derivative of $\ogw_{lb}$ w.r.t. $\Cvec$ accordingly.
  Afterwards, considering the barycenter problem,
  we will update the cost matrix of the barycenter by the weighted summation over the dissimilarity between sample and center.
  To be specific,
  \begin{align}
    \frac{\partial \Bcal (\Cvec)}{\partial \Cvec}
      = \sum_i \lambda_i \frac{\partial \ogw(\Cvec, \Dvec_i)}{\partial \Cvec}
      = \sum_i \lambda_i \sbr{\frac{2\Cvec}{m^2} - \frac{2}{mn} \frac{\partial \tr (\Cvec \Pvec \Dvec_i \Pvec^\top)}{\partial \Cvec}}.
  \end{align}
  Therefore, we can plug the partial derivative of the trace in terms of lower and upper bounds accordingly.

 \subsection{Dataset statistics}
  \label{subsec:app_dataset_stats}
  Table~\ref{tab:datasets} provides the statistics of the six datasets used in graph classification experiment.
  \begin{table*}
    \centering
    \caption{Statistics of the datasets}
    \label{tab:datasets}
    \resizebox{\textwidth}{!}{%
      \begin{tabular}{l*{9}c}
        \toprule
        dataset & \# graphs & \# class & \# features & ave. edge & min edge & max edge & avg. node & min node & max node \\
        \hline
        BZR     & 405       & 2        & 56          & 74.0      & 26       & 120      & 35.0      & 13       & 57       \\
        COX2    & 467       & 2        & 38          & 86.0      & 68       & 118      & 41.0      & 32       & 56       \\
        MUTAG   & 188       & 2        & 7           & 38        & 20       & 66       & 17.5      & 10       & 28       \\
        PTC-MR  & 344       & 2        & 18          & 25.0      & 2        & 142      & 13.0      & 2        & 64       \\
        IMDB-B  & 1000      & 2        & -           & 130       & 52       & 2498     & 17        & 12       & 136      \\
        IMDB-M  & 1500      & 3        & -           & 72        & 23       & 2934     & 10        & 7        & 89       \\
        \bottomrule
      \end{tabular}}
  \end{table*}

 \subsection{Supplementary Measures of Tightness}
  \label{subsec:app_tightness}
\begin{figure}[h]
    \centering
    \subfloat[Gap on BZR]{
      \includegraphics[width=0.3\textwidth]{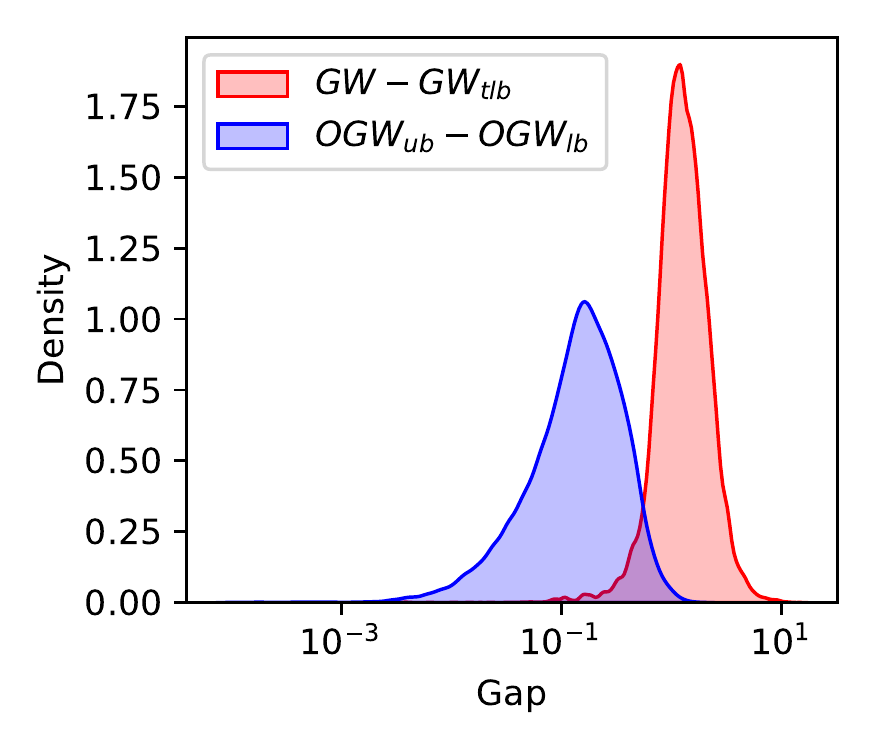}
      \label{fig:tightness_BZR}
    }
    \subfloat[Gap in OGW and its variants]{
      \includegraphics[width=0.27\textwidth]{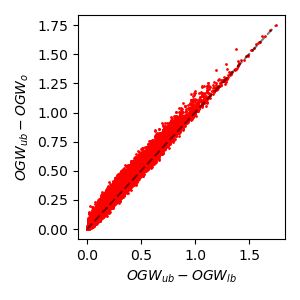}
      \label{fig:tightness_ogw_lbs_BZR}
    }
    \caption{Tightness of lower and upper bounds for $\gw$ and $\ogw$ (BZR dataset)}
    \label{fig:app_BZR}
  \end{figure}

\begin{figure}[h]
    \centering
    \subfloat[Gap on COX2]{
      \includegraphics[width=0.3\textwidth]{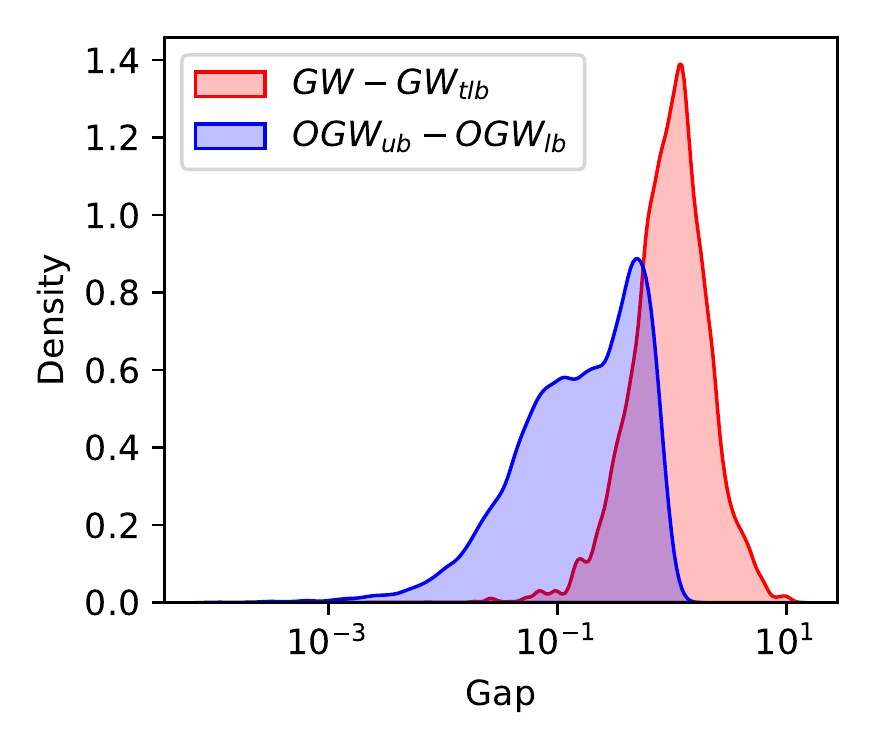}
      \label{fig:tightness_COX2}
    }
    \subfloat[Gap in OGW and its variants]{
      \includegraphics[width=0.27\textwidth]{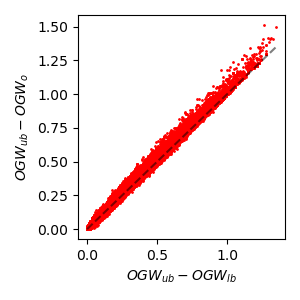}
      \label{fig:tightness_ogw_lbs_COX2}
    }
    \caption{Tightness of lower and upper bounds for $\gw$ and $\ogw$ (COX2 dataset)}
  \end{figure}

\begin{figure}[h]
    \centering
    \subfloat[Gap on PTC\_MR]{
      \includegraphics[width=0.3\textwidth]{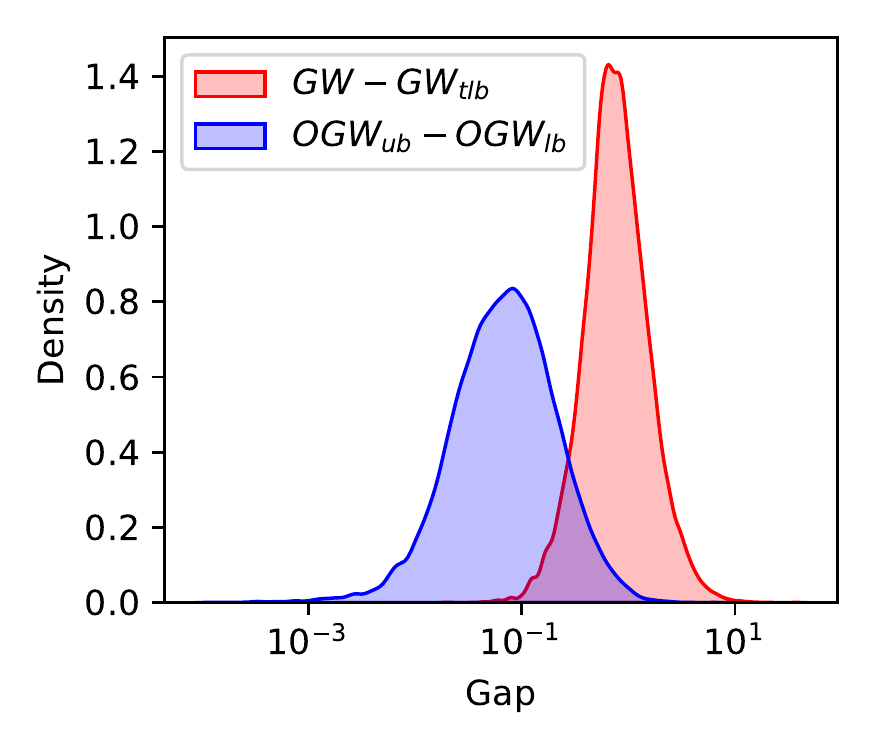}
      \label{fig:tightness_PTC_MR}
    }
    \subfloat[Gap in OGW and its variants]{
      \includegraphics[width=0.27\textwidth]{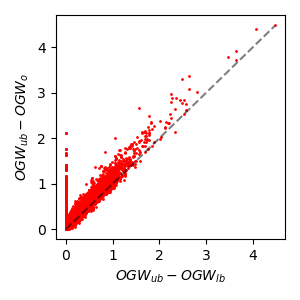}
      \label{fig:tightness_ogw_lbs_PTC_MR}
    }
    \caption{Tightness of lower and upper bounds for $\gw$ and $\ogw$ (PTC\_MR dataset)}
  \end{figure}

\begin{figure}[h]
    \centering
    \subfloat[Gap on IMDB-BINARY]{
      \includegraphics[width=0.3\textwidth]{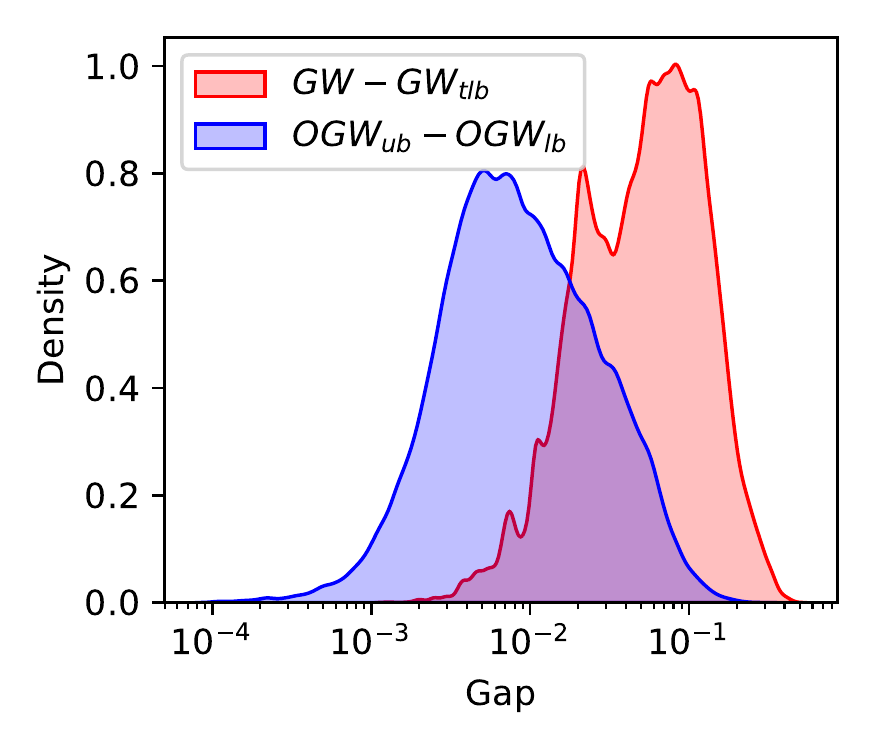}
      \label{fig:tightness_IMDB-BINARY}
    }
    \subfloat[Gap in OGW and its variants]{
      \includegraphics[width=0.27\textwidth]{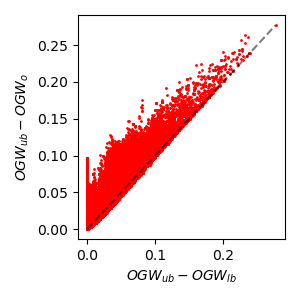}
      \label{fig:tightness_ogw_lbs_IMDB-BINARY}
    }
    \caption{Tightness of lower and upper bounds for $\gw$ and $\ogw$ (IMDB-BINARY dataset)}
  \end{figure}

\begin{figure}[h]
    \centering
    \subfloat[Gap on IMDB-MULTI]{
      \includegraphics[width=0.3\textwidth]{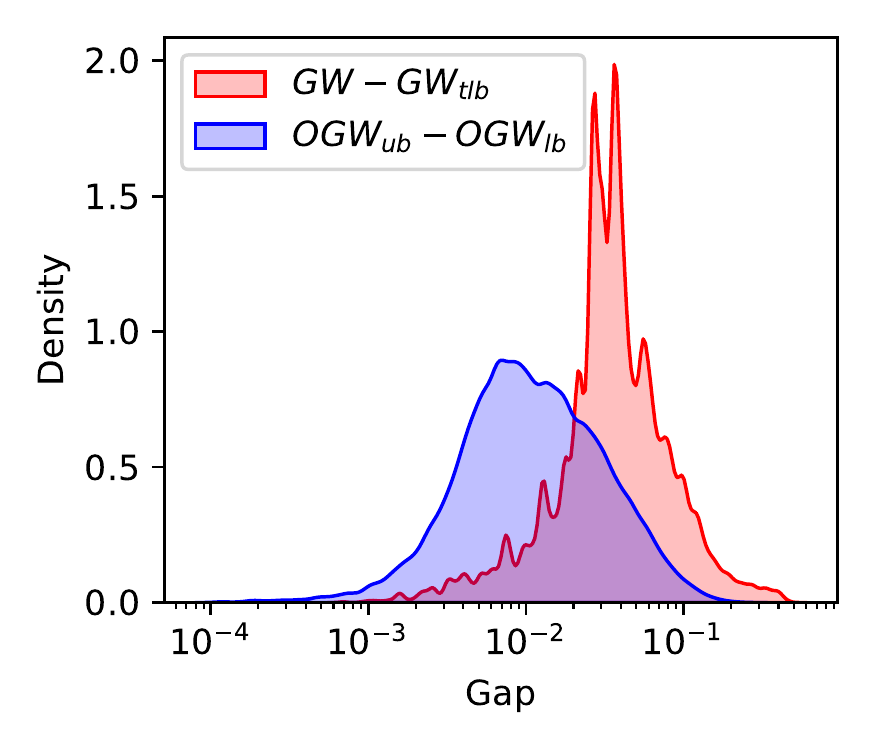}
      \label{fig:tightness_IMDB-MULTI}
    }
    \subfloat[Gap in OGW and its variants]{
      \includegraphics[width=0.27\textwidth]{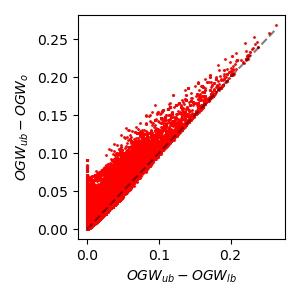}
      \label{fig:tightness_ogw_lbs_IMDB-MULTI}
    }
    \caption{Tightness of lower and upper bounds for $\gw$ and $\ogw$ (IMDB-MULTI dataset)}
    \label{fig:app_IMDB-M}
  \end{figure}

  Similar to the tightness measures for MUTAG, we also provide the tightness for datasets BZR, COX2, PTC-MR, IMDB-B, IMDB-M from Figure~\ref{fig:app_BZR} to \ref{fig:app_IMDB-M}.
  They confirm our observation that our lower bound of $\ogw$ approximates it significantly more tightly than the third lower bound does for $\gw$.

 \subsection{Samples for the Barycenter Problem}
  \label{subsec:app_bary_samples}

\begin{figure}
    \centering
    \includegraphics[width=0.5\textwidth]{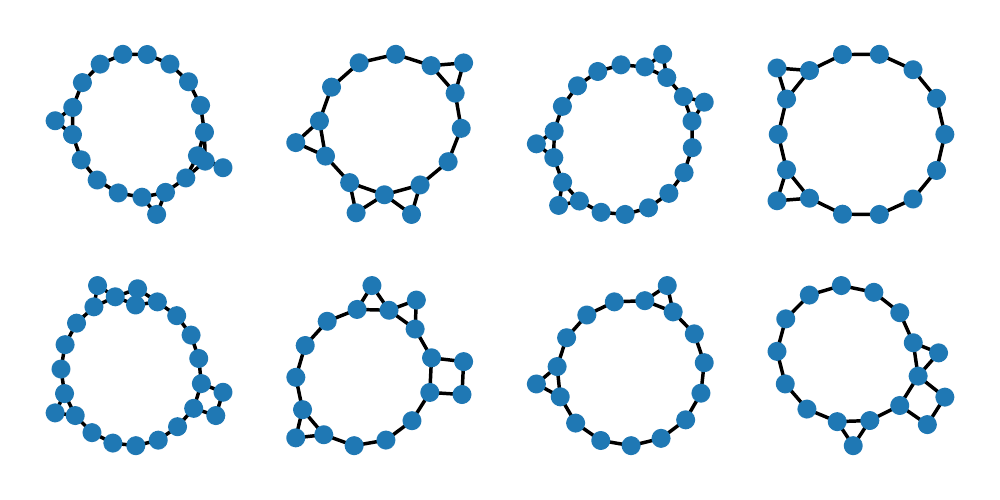}
    \caption{Generated sample cycle-like graphs with structural noises.}
    \label{fig:barycenter_examples}
  \end{figure}

\begin{figure}
    \centering
    \includegraphics[width=0.75\textwidth]{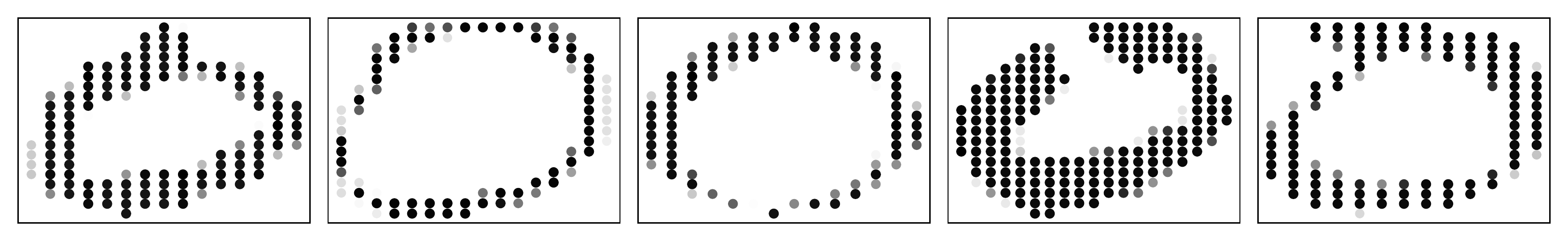}
    \includegraphics[width=0.75\textwidth]{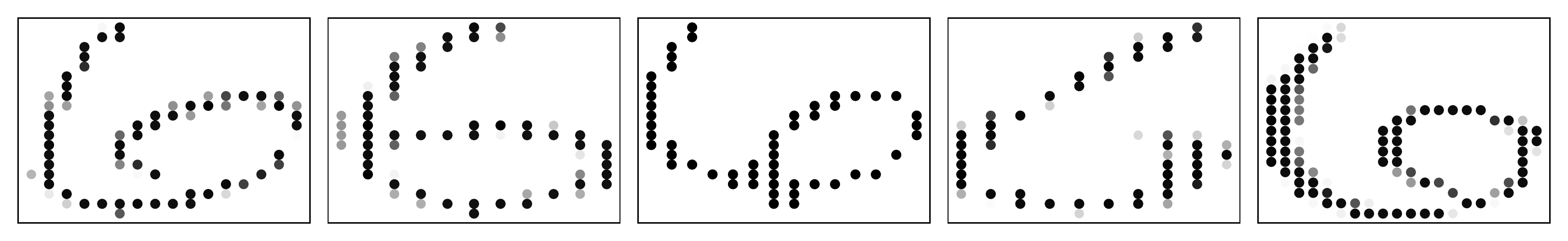}
    \includegraphics[width=0.75\textwidth]{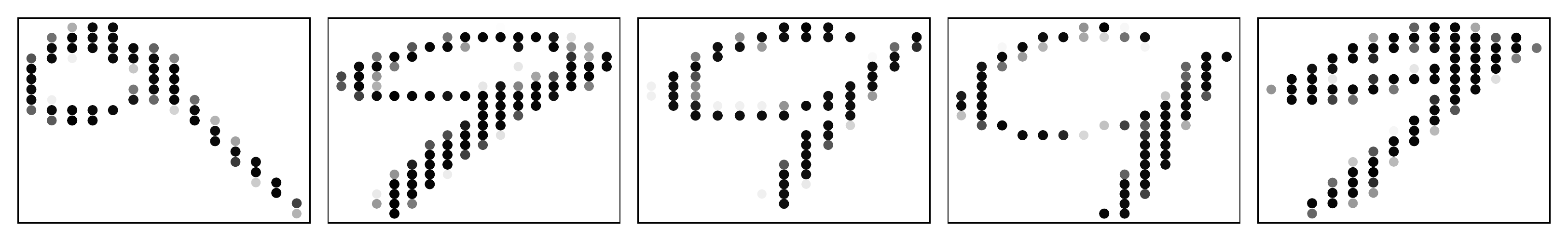}
    \caption{Samples from point cloud MNIST-2D}
    \label{fig:barycenter_examples_mnist}
  \end{figure}

  Figure~\ref{fig:barycenter_examples} are the synthetic noised graphs generated by randomly adding structural noises.
  And Figure~\ref{fig:barycenter_examples_mnist} are the real samples from point cloud MNIST dataset,
  and the rows correspond to labels 0, 6, and 9.

  %  \appendix
  %  % NOTE: necessary when ptmx or no mathfont class option is given
  %  \providecommand{\upGamma}{\Gamma}
  %  \providecommand{\uppi}{\pi}
  % \section{Math font exposition}
  %  How math looks in equations is important:
  %  \begin{equation*}
  %    F_{\alpha,\beta}^\eta(z) = \upGamma(\tfrac{3}{2}) \prod_{\ell=1}^\infty\eta \frac{z^\ell}{\ell} + \frac{1}{2\uppi}\int_{-\infty}^z\alpha \sum_{k=1}^\infty x^{\beta k}\mathrm{d}x.
  %  \end{equation*}
  %  However, one should not ignore how well math mixes with text:
  %  The frobble function \(f\) transforms zabbies \(z\) into yannies \(y\).
  %  It is a polynomial \(f(z)=\alpha z + \beta z^2\), where \(-n<\alpha<\beta/n\leq\gamma\), with \(\gamma\) a positive real number.

\end{document}

%% file: fig_qap_connection.tex
\begin{figure}
  \centering
  \begin{tikzpicture}
    % Nodes
    \node[align=center]  (qap) [text width=1.5cm] at (0, 0)
    {QAP \scriptsize $\Ecal \cap \Ncal \cap \Ocal$};
    \node[align=center]  (gw) [text width=1cm] at (-2.5, 0)
    {GW  \scriptsize $\Ecal \cap \Ncal$};
    \node[align=center]  (ogw) [text width=1cm] at (2.5, 0)
    {$\ogw$ \scriptsize $\Ecal \cap \Ocal$};
    % Arrays
    \draw[->] (qap) -- (gw);
    \draw[->] (qap) -- (ogw);
  \end{tikzpicture}
  \caption{Connection between QAP and GW, $\ogw$}
  \label{fig:conn_qap_gw_ogw}
\end{figure}

%% file: fig_transition.tex
\begin{figure}
  \centering
  \begin{tikzcd}[column sep=width("clipping at 0")]
    \gw
    \arrow[d, "\text{add node}"{sloped}, "\text{feature}"'{sloped} ]
    \arrow[r, "\text{change~} \Ncal \cap \Ecal", "\text{into~} \Ocal \cap \Ecal"' ]
    &
    \ogw
    \arrow[d, "\text{add node}"{sloped}, "\text{feature}"'{sloped}]
    \\
    [4ex]
    \fgw
    \arrow[r, "\text{change~} \Ncal \cap \Ecal", "\text{into~} \Ocal \cap \Ecal"']
    &
    \ofgw
  \end{tikzcd}
  \caption{Steps of constructing the graph discrepancies}
  \label{fig:transition}
\end{figure}